\documentclass[sigconf]{acmart}
\AtBeginDocument{%
  \providecommand\BibTeX{{%
    \normalfont B\kern-0.5em{\scshape i\kern-0.25em b}\kern-0.8em\TeX}}}

\setcopyright{acmlicensed}
\copyrightyear{2018}
\acmYear{2018}
\acmDOI{XXXXXXX.XXXXXXX}

\acmConference[Conference acronym 'XX]{Make sure to enter the correct
  conference title from your rights confirmation emai}{June 03--05,
  2018}{Woodstock, NY}
\acmISBN{978-1-4503-XXXX-X/18/06}

\usepackage{graphicx}
\usepackage{pifont}
\usepackage{textcomp}
\usepackage{xcolor}
\usepackage{algorithm}
\usepackage{algpseudocode}
\usepackage{amsmath}
\usepackage{newfloat}
\usepackage{listings}
\usepackage[caption=false,font=normalsize,labelfont=sf,textfont=sf]{subfig}
\usepackage{tikz}
\usepackage{varwidth}
\usepackage{multirow} 
\usepackage{hyperref}

\begin{document}

% \title{Can diffusion model-based reconstruction help Out-of-distribution detection on molecular graphs}
% \title{Revolutionizing Molecular Graph OOD Detection with a Novel Diffusion Model: Unveiling a Scalable and Efficient Approach}

\title{Optimizing OOD Detection in Molecular Graphs: A Novel Approach with Diffusion Models}

\author{Xu Shen}
\authornote{Both authors contributed equally to this research.}

\affiliation{%
  \institution{Jilin University}
 \streetaddress{}
  \city{Changchun}
  \country{China}}
  \email{xushen23@mails.jlu.edu.cn}

\author{Yili Wang}
\authornotemark[1]
\affiliation{%
  \institution{Jilin University}
  \streetaddress{}
  \city{Changchun}
  \country{China}}
\email{wangyl21@mails.jlu.edu.cn}

\author{Kaixiong Zhou}
\affiliation{%
  \institution{Massachusetts Institute of Technology}
  \streetaddress{}
  \city{Cambridge}
  \country{USA}}
\email{kz34@mit.edu}

\author{Shirui Pan}
\affiliation{%
  \institution{Griffith University}
  \streetaddress{}
  \city{Goldcoast}
  \country{Australia}}
\email{s.pan@griffith.edu.au}

\author{Xin Wang}
\authornote{Corresponding author.}
\affiliation{%
  \institution{Jilin University}
  \streetaddress{}
  \city{Changchun}
  \country{China}}
\email{xinwang@jlu.edu.cn}


% \affiliation{%

\renewcommand{\shortauthors}{Trovato and Tobin, et al.}

\begin{abstract}
Despite the recent progress of molecular representation learning, its effectiveness is assumed on the close-world assumptions that training and testing graphs are from identical distribution. The open-world test dataset is often mixed with out-of-distribution (OOD) samples, where the deployed models will struggle to make accurate predictions. The misleading estimations of molecules' properties in drug screening or design can result in the tremendous waste of wet-lab resources and delay the discovery of novel therapies. Traditional detection methods need to trade off OOD detection and in-distribution (ID) classification performance since they share the same representation learning model. In this work, we propose to detect OOD molecules by adopting an auxiliary diffusion model-based framework, which compares similarities between input molecules and reconstructed graphs. Due to the generative bias towards reconstructing ID training samples, the similarity scores of OOD molecules will be much lower to facilitate detection. Although it is conceptually simple, extending this vanilla framework to practical detection applications is still limited by two significant challenges. First, the popular similarity metrics based on Euclidian distance fail to consider the complex graph structure. Second, the generative model involving iterative denoising steps is notoriously time-consuming especially when it runs on the enormous pool of drugs. To address these challenges, our research pioneers an approach of \textbf{P}rototypical \textbf{G}raph \textbf{R}econstruction for \textbf{M}olecular \textbf{OO}d \textbf{D}etection, dubbed as PGR-MOOD. Specifically, PGR-MOOD hinges on three innovations: i) An effective metric to comprehensively quantify the matching degree of input and reconstructed molecules according to their discrete edges and continuous node features; ii) A creative graph generator to construct a list of prototypical graphs that are in line with ID distribution but away from OOD one; iii) An efficient and scalable OOD detector to compare the similarity between test samples and pre-constructed prototypical graphs and omit the generative process on every new molecule. Extensive experiments on ten benchmark datasets and six baselines are conducted to demonstrate our superiority: PGR-MOOD achieves more than $8\%$ of average improvement in terms of detection AUC and AUPR accompanied by the reduced cost of testing time and memory consumption. The
anonymous code is in: \href{https://anonymous.4open.science/r/PGR-MOOD-53B3}{https://anonymous.4open.science/r/PGR-MOOD-53B3}.

\end{abstract}





\maketitle

\section{Introduction}
Molecular representation learning, which transforms molecules into low-dimensional vectors, has emerged as a critical and essential part of many biochemical problems, such as drug property prediction~\cite{molecular, GraphSAM} and drug design~\cite{drugdesign}. For handling the non-Euclidean molecules, graph neural networks (GNNs) have been widely applied to encode both node features and structural information based on message-passing strategy~\cite{mp}. The embedding vectors of atoms and/or edges are then summarized to represent the underlying molecules and adopted to various downstream tasks~\cite{graphsage, GIN,linegraph}.

\begin{figure}[t!]   
					\centering          
					\subfloat[ID and OOD molecules]   
					{
						\label{fig1:1}\includegraphics[width=0.21\textwidth]{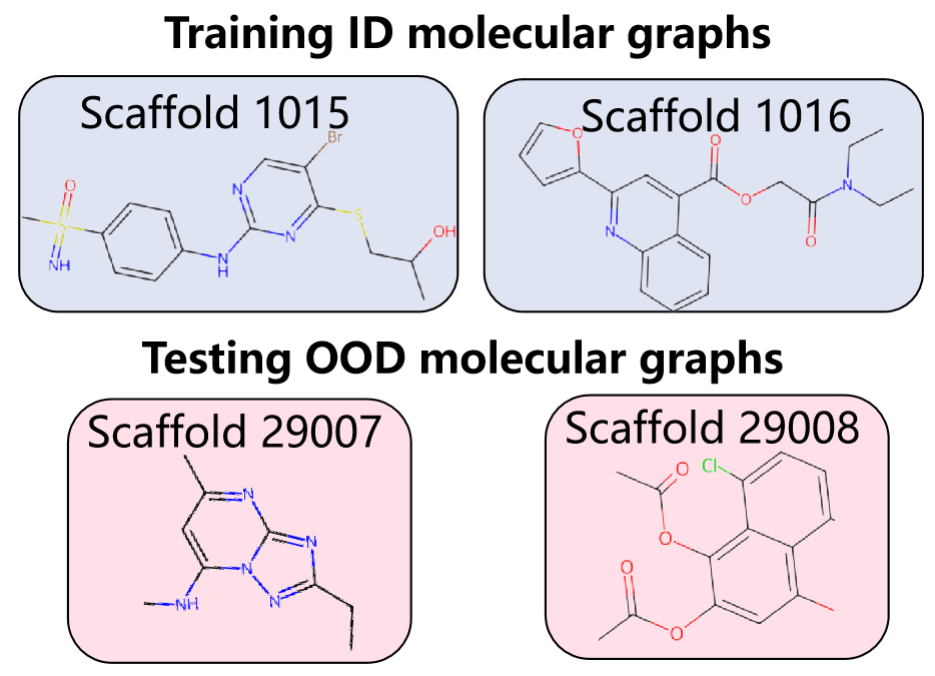}
						
					}\hspace{1mm}
					\subfloat[Loss and auroc]
					{
						\label{fig1:2}\includegraphics[width=0.23\textwidth,height=0.13\textheight]{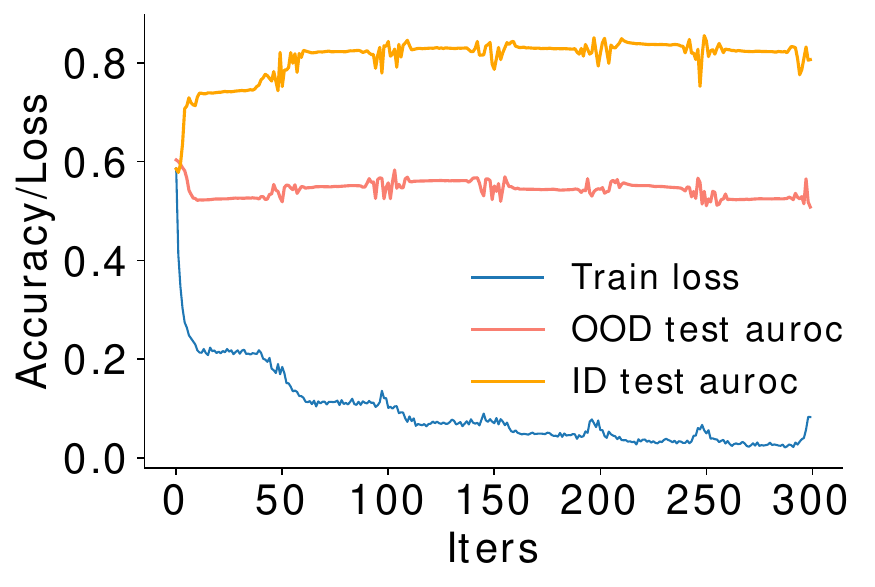}
					}
					
					\vspace{-5pt}
					\caption{(a) Illustration of OOD and ID molecules, which have different scaffolds or sizes, or both. (b)  Vanilla GCN's performance declines rapidly when testing on OOD graphs, even though it performs well on ID graphs. }
					\label{fig:ob1}          
     \vspace{-5pt}
\end{figure}

The recent successes of molecular representation learning are often built on the assumption that training and testing graphs are from identical distribution. However, out-of-distribution (OOD) molecular graphs with different scaffolds or sizes, as shown in Fig. \ref{fig1:1}, is unavoidable when the model is deployed in real-world scenarios~\cite{drugood}. Taking antibiotics screening as example, the training data consists of drugs inhibiting the growth of Gram-negative pathogens, while the testing data is mixed with antibiotics against Gram-positive ones~\cite{liu2023deep}. Because of the different pharmacological mechanisms in treating bacteria, a reliable drug screening model should not only accurately identify more the in-distribution (ID) samples (e.g., Gram-negative), but also detect ``unknown'' OOD inputs (e.g., Gram-positive) to avoid misleading predictions during inference. As illustrated in Fig.~\ref{fig1:2}, a notable decline in GNNs' prediction accuracy is observed with OOD samples. This highlights the significance of OOD detection, which discerns between ID and OOD inputs, allowing the model to adopt appropriate precautions~\cite{msp}.

Prior arts of graph OOD detection can be roughly grouped into two categories. One line of the existing work aims to leverage the original classifier and fine-tune it to improve its detection ability~\cite{graphde, Good-d}. The another line is to redesign the scoring function to indicate ID and OOD cases~\cite{AAGOD,graphsafe}. 
% One line of the OOD detection methods aims to leverage the original classifier and fine-tune it to improve its detection ability~\cite{graphde, Good-d}. The other method is to redesign the scoring function to indicate ID and OOD cases~\cite{AAGOD,graphsafe}. 
Nevertheless, these methods inevitably require modifications to the original molecular representation learning model, leading to a trade-off between OOD detection and ID prediction~\cite{DiffGuard}. 
Recent advancements in computer vision %(CV) area 
have proposed the use of a diffusion model-based reconstruction approach for the unsupervised OOD detection, which typically involves an auxiliary generative model that approximates the ID distribution to reconstruct the input samples during testing phase~\cite{diffusionOOD,DiffGuard,LMD}. 
% Since the quality of the reconstruction belonging to ID is much higher than that of OOD, 
Since the distribution of reconstructed samples is more biased towards ID than OOD, the disparity between original inputs and reconstructed outputs can be used as a judge metric for OOD detection. However, this kind of approach has never been practiced in the field of molecular graphs.

\begin{figure}[!t]
		\centerline{\includegraphics[scale=0.47]{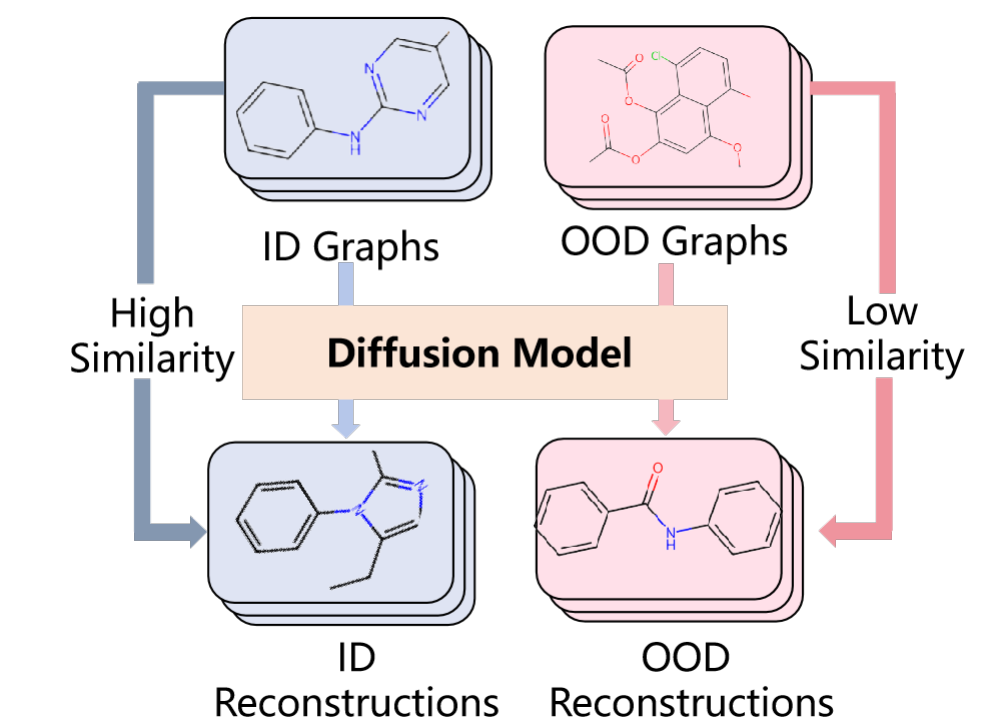}}
  \vspace{-5pt}
		\caption{	Illustration of reconstruction-based OOD detection with the diffusion model. ID and OOD share different similarities with their respective reconstruction graphs and can be used as a score for OOD detection.		}	\label{reconstructionfig}
  \vspace{-15pt}
	\end{figure}

We first design a naive model called GR-MOOD as shown in Fig. \ref{reconstructionfig}, to verify the feasibility of the reconstruction method for molecular  OOD detection and draw a positive conclusion through experiments. %The Diffusion models, while demonstrating promising outcomes in OOD detection within the computer vision (CV) domain~\cite{LMD}, encounter distinct challenges when adapted to graph data analysis
However, the inherent complexity of molecular graphs, which are characterized by non-Euclidean structures, poses two significant challenges. 
% In the first place, 
First, this nature of molecular graphs renders conventional similarity metrics (e.g., Euclidean distance) less effective to quantify the closeness between original and reconstructed graphs.  Meanwhile, the different molecules often undergo distribution shifts that include both structural and feature changes, further complicating the assessment of similarity. This leads to \textit{Challenge 1: Identifying an effective metric to evaluate the similarity between the original input and the reconstruction.}
More importantly, the diffusion models require hundreds or thousands of sampling steps to denoise from a normal standard distribution towards generating new graphs, which introduces additional complexity. 
% involving iterative noise reduction from a normal standard distribution and hundreds or thousands of sampling steps is required in this generation process which introduces additional complexity. 
Such extensive requirement becomes impractical, especially when performing reconstructing for a large volume of test samples. This leads to \textit{Challenge 2: Addressing the additional complexity of diffusion model required for reconstruction.} 
Thus we propose a critical research question: 
\noindent\colorbox{lightgray}{\begin{varwidth}{0.98\linewidth}\textit{How can we adopt reconstruction method to effectively and efficiently handle the unique properties of molecular graphs for OOD detection?}\end{varwidth}}

% To tackle these challenges, 
In this paper, we introduce a groundbreaking OOD detection model, \textbf{P}rototypical \textbf{G}raph \textbf{R}econstruction for \textbf{M}olecular \textbf{OO}d \textbf{D}etection (PGR-MOOD for short). 
For Challenge 1, concerning the identification of an effective metric for assessing the similarity between the original input and its reconstruction, %PGR-MOOD shifts focus to the degree of matching between graphs. We utilize the Fused Gromov-Wasserstein (FGW) distance as a sophisticated metric to quantify similarity in graph reconstructions. This choice not only fully utilizes the structural and feature information of molecular graphs, but also enhances the precision of similarity evaluation \textcolor{red}{[KX: not clear why and how it can enhance precision in high level]}.
PGR-MOOD adopts Fused Gromov-Wasserstein (FGW) distance~\cite{FGW}, which utilizes both the structural and feature information of molecular graphs to enhance the measurement of their matching degree. 
% is the optimal measurement to directly calculate the matching degree between graphs, and can utilize both the structural and feature information of molecular graphs.
% This choice not only enhances the precision of similarity evaluation but also streamlines the OOD detection process. 
% Instead, it employs a diffusion model during the training phase to generate prototypical graphs that embody the characteristics of all reconstructions. 
% This strategic move allows for the direct comparison of these prototype graphs with test samples in the detection phase, this initiative has significantly reduced the testing time of the diffusion model. 
% Additionally, we utilize the Fused Gromov-Wasserstein (FGW) distance as a sophisticated metric to quantify similarity in prototypical reconstructions. This choice not only enhances the precision of similarity evaluation but also streamlines the OOD detection process.
To efficiently address Challenge 2, PGR-MOOD proposes  to create a series of prototypical graphs that are closer to ID samples and away from OOD ones. We reduce the need of reconstructing every test graph and just compare its similarities with the prepared prototypical graphs. With this procedure, we can extend to the large-scale OOD detection. Our contributions are summarized as follows:
\begin{itemize}
    \item \textbf{GR-MOOD Framework:} We propose to detect OOD graphs from a novel perspective, i.e., via comparing the original molecules with their reconstructed outputs based on the diffusion model. The technical feasibility and challenges are analyzed empirically for this new framework. 
    % The first attempt to verify whether OOD detection of molecular graphs based on reconstruction is feasible. We draw positive conclusions through experiments and summarize two challenges regarding performance and efficiency. 

    \item \textbf{PGR-MOOD Framework:} To overcome the  challenges of reconstruction measurement and generation efficiency, we propose a molecular detection method that contains a prototypical graphs generator and a similarity function based on FGW distance. In the testing phase, one only needs to measure the similarity between the prototypical graphs and the current inputs to identify OOD with lower values.

     % a novel method that leverages diffusion models for accurate out-of-distribution (OOD) detection in molecular graphs by transforming classification tasks into similarity assessments.
    
    %\item \textbf{Prototypical Graph Generation:} Simplifies OOD detection with prototypical graph generation by a unique loss function. Substantial savings in time and memory consumption than the GR-MOOD and the existing SOTA.
    % enhancing model distinction between ID and OOD samples.
    % PGR-MOOD employs a novel strategy of generating prototypical graphs during training, facilitated by a specifically designed loss function. 
    \item \textbf{SOTA Experimental Results:} We conduct extensive analysis on ten benchmark molecule datasets and compare with six baselines. PGR-MOOD obtains the consistent superiority over other state-of-the-art models, delivering the average improvements of AUC and AUPR by $8.54\%$ and $8.15\%$, $13.7\%$ reduction on FPR95, and substantial savings in time and memory consumption. % It has been rigorously tested across ten diverse molecular graph datasets.
\end{itemize}

\section{related work}
\subsection{Graph Neural Networks}
Since graph neural networks can use the topological structure and node properties of graphs for representation learning, they have become the most powerful method for processing graph data~\cite{KP-GNN, GNN-AK, ESAN,zhou2020towards}, especially molecular graphs~\cite{mclGNN,drugGNN}. 
GCN~\cite{GCN}, the simplest but most efficient method, has been proved to be equivalent to the first-order approximation filter on graphs~\cite{Bernnet} and thus performs well in node classification~\cite{graphsage} and link prediction~\cite{linegraph}.
On graph instance-related tasks, GIN~\cite{GIN} proves that GNN is as powerful as the 1-WL test and leverages an injective summation operation to increase performance. 
% To enhance the representation ability, ESAN~\cite{ESAN} represents each graph as a set of subgraphs and uses a suitable equivariant architecture to process them. 
% KP-GNN~\cite{KP-GNN} uses diffusion kernel and shortest path kernel to enlarge the aggregation radius of nodes and capture more neighborhoods once. 
% GNN-AK~\cite{GNN-AK} extends local aggregation to general subgraph schema to improve expressiveness while ensuring scalability. 
More and more researchers have proposed more representational methods, but they all ignore the performance and trustworthiness issues brought by OOD distribution~\cite{trustworthy,trustworthysurvey}.

\subsection{Graph Generative Models}
Graph generative models aim to learn the distribution of the graph data and sample from it to generate novel graphs~\cite{generationsurvey}, especially for molecular graphs since it is related to many science issues~\cite{drugdiscovery, materialsgraph, protein}. 
Some graph generation methods are inspired by auto-regressive models, such as VAE-based~\cite{VAE} or normalizing flow-based models~\cite{molgrow}. 
However, they are limited by the high computational cost and inability to model permutation invariance of graph~\cite{gdss}. 
Inspired by the diffusion models in computer vision~\cite{diffusionmodel}, the same insight on graphs has developed in recent years~\cite{edpgnn, digress, EDGE}. 
% EDP-GNN utilizes score matching to build graph data distribution~\cite{edpgnn}. 
% GDSS models the joint distribution of the nodes and edges through a system of stochastic differential equations (SDEs)~\cite{gdss}. 
% DiGress changes the generation process from continuous to discrete, which is more suit for the graph data~\cite{digress}. EDGE aims to balance the scalability and efficiency of the generation~\cite{EDGE}.
Although diffusion models achieve state-of-the-art performance, they still suffer from inefficiencies caused by slow denoising processes~\cite{sagess}.

\subsection{OOD Detection on Graphs}
Recently, many studies focus on graph OOD detection due to its importance.
GOOD-D is the pioneering work for unsupervised OOD graph detection, which performs hierarchical contrastive learning to capture latent ID patterns and detects OOD graphs based on their semantic inconsistency~\cite{Good-d}. %However, these approaches need to train a GNN specialized for OOD detection from scratch, which may have high computation cost [19]. Differently,
GraphDE determines ID and OOD by inferring the environment variables of the graph generation process~\cite{graphde}.
AAGOD aims to learn a  parameterized amplifier matrix to emphasize the key patterns which helpful for graph OOD detection, thereby enlarging the gap between OOD and ID graphs~\cite{AAGOD}. 
Anomaly graph detection can also be seen as a special case of OOD detection, since anomaly graphs with anomaly structures and features can be caused by distribution shifts and many methods have been proposed to solve it~\cite{OCGIN, GLocalKD}. 
% OCGIN redesigns and trains GNNs as a binary classifier to distinguish between ID and OOD~\cite{OCGIN}. GLocalKD learns graph global and local pattern information as the basis for distinguishing ID and OOD by joint distillation of graph and node representations~\cite{GLocalKD}. 
All of the above methods require redesigning or training well-performing GNNs on the ID datasets and inevitably lead to a trade-off between OOD detection and ID prediction.

\section{Preliminaries }
% In this section, we introduce notation and formalize the out-of-distribution detection, graph neural networks, and graph generation model, respectively. \textcolor{red}{KX: Do not need this paragraph.}

We define an undirected graph ${G} = \left ( A, X \right ) $ with $n$ nodes, where %$\left | V \right | =n$ is the number of nodes, $\left | E \right | =m$ is the number of edges. We define
$A \in \mathbb{R}^{n\times n}$ is adjacency matrix to represent the graph topology, $X \in \mathbb{R}^{n\times d}$ is feature matrix of all nodes with the dimensionality of $d$. $G$ can also be re-written by Optimal transmission (OT) format~\cite{TGNN} to represent as a tuple $({A},{X},{\mu})$, where $\mu\in\mathbb{R}^{n}$ is a vector of weights modeling the relative importance of the
nodes and we define it as a uniform weight $(\mathbf{1}_{n}/n)$. In addition, we define $D_{\mathrm{train}}$ %\textcolor{red}{[KX: underscipt changes to $D_{\mathrm{train}}$, the same in the folloing]} 
as the training dataset that usually consists of ID graphs, and define $D_{\mathrm{test}}$ as the test dataset, which can be divided into in-distribution subset $D^{\mathrm{in}}_{\mathrm{test}}$ and out of distribution subset $D^{\mathrm{out}}_{\mathrm{test}}$. 
% \SP{Should be $D^{\mathrm{out}}_{\mathrm{test}}$?} %Each graph $G_{test}$ in $D_{\mathrm{test}}$ corresponds to a $y_{test}$ that needs to be predicted, which is a label vector for classification task and a scalar for regression task.  

%	\begin{figure}[htbp]
%		\centerline{\includegraphics[scale=0.9]{fgw.png}}
%		\caption{$\operatorname{FGW}$ for the optimal coupling $\pi^{\ast }$ on the similarity between different node feature $(A_{ij}-\overline{A}_{kl})^2$ and structure similarities $|{X}_i-\overline{{X}}_k\|_2^2$.
 % }	\label{smoothfig}
	%\end{figure}
\subsection{Out of Distribution Detection }
For OOD detection task, we aim to design a detector $g$ to distinguish whether the input graph $G$ is an OOD sample or not:
\begin{align}\label{OODdetect}
    \left.g(G;\tau,J)=\left\{\begin{array}{ll}0\ \text{(OOD)},&\text{if}\ J(G)\leq\tau,\\1\ \text{(ID)},&\text{if}\ J(G)>\tau.\end{array}\right.\right.
\end{align}
%where $\tau$ is the threshold for the judging function $J$ to identify the OOD graph.
where $J$ denotes a judging function to score the input molecules and $\tau$ denotes threshold for identifying the OOD samples.
A desired OOD detector should assign judge scores with the maximum gap between ID and OOD samples.
%A good OOD detector should enlarge the judge score gap between ID graphs and OOD graphs
This target can be described as the following optimization:
\begin{align}\label{OODoptimize}
    \max_{J}\mathbb{E}_{G\sim D^{\mathrm{in}}_{\mathrm{test}}}J(G)-\mathbb{E}_{G\sim D^{\mathrm{out}}_{\mathrm{test}} }J(G).
\end{align}
Supposing the judge score distributions of ID and OOD have significant divergence, we can distinguish them with a simple intermediate threshold. For reconstruction-based OOD detection as shown in Fig.~\ref{reconstructionfig}, the similarity between the input and the output molecules of diffusion model $\mathrm{F_M}$ is often adopted as the judge function: 
\begin{align} \label{judgesim}
    J(G) &= \operatorname{sim}(\mathrm{F_M}(G),G),
\end{align}
where $\mathrm{F_M}(G)$ is the reconstructed output and $\operatorname{sim}(\cdot)$ is the similarity function. OOD inputs correspond to the lower reconstruction quality and therefore the lower similarity, while the similarity measurement is higher for the ID inputs.

\subsection{Graph Neural Networks}
The typical GNNs are based on message passing paradigm. Specifically, the final representation of graph $G$ for a $L$-layer GNNs is:
\begin{align}
		\label{message-passing} m_{v} ^{(L)} &= \operatorname{MP}\left (m_{v} ^{(L-1)},\{ (m_{u}^{ (L-1)}) ,u\in N\left ( v \right )   \}\right ) ,\\ 
		\label{pooling}	z_{G} &=  \operatorname{Pooling}\left(\left\{{m}_v^{(L)}\mid v\in G\right\}\right),
			%	\end{aligned}
	\end{align}
where $m^{(0)}_{v}=X_{v}$ is raw node feature, $N(v)$ represents a set of neighbor nodes with respect to node $v$, and $\operatorname{MP}$ is the message passing process that aggregates neighborhood features (e.g., sum, mean, or max) and combines them with the local node. GNNs iteratively perform $\operatorname{MP}$ to learn the effective node representations and utilize function $\operatorname{Pooling}$ to map all the node representations into the graph representations, which is a single vector. 

\subsection{Graph Generative Model}
%Diffusion model is one of  state-of-the-art 
The generative method based on the diffusion model consists of a forward diffusion process and a reverse denoising process. At the forward process, the model progressively adds noise to the original data until a standard normal distribution. At the reverse process, the model learns the score function (i.e., a neural network) to remove the perturbed noise with the same amount of steps~\cite{surveydiffusion,DCT,diffusionmodel}. 

Given a graph ${G}=(A,X)$, we can use continuous time $t\in[0,T]$ to index the diffusion trajectory $\left \{{G_{t}}=(A_{t},X_{t})  \right \}^{T}_{t=1}$, such that $G_{0}$ is the original input graph and $G_{T}$ approximately follows the normal distribution. The forward process transforms $G_{0}$ to $G_{T}$ through a stochastic differential equation (SDE):
\begin{align}\label{addnoise}
\mathrm{d}{G}_t=\mathbf{f}_t({G}_t)\mathrm{d}t+g(t)\mathrm{d}\mathbf{w},
\end{align}
where $\mathbf{w}$ is standard Wiener process \cite{gdss}, $\mathbf{f}_t(\cdot): \mathcal{G}\to\mathcal{G}$ is linear drift coefficient, $g(t): \mathbb{R} \to \mathbb{R}$ is a scalar function which represents the diffusion coefficient. $\mathbf{f}_t({G}_t)$ and $g(t)$ relate to the amount of noise $\mathrm{d}\mathbf{w}$ added to the graph at each infinitesimal step $\mathrm{d}t$. In order to generate graphs that follow the distribution of $G_{0}$, we start from $G_{T}$ and utilize a reverse-time SDE %with scores of perturbed graphs $\nabla_{G_t}\log p_t(G_t)$
for denoising from $T$ to $0$:
\begin{align}\label{reverse}
    \mathrm{d}G_t=\left[\mathbf{f}_t(G_t)-g(t)^2 \mathrm{S}_{\theta}(G_t,t)\right]\mathrm{d}t+g(t)\mathrm{d}\bar{\mathbf{w}},
\end{align}
where $\mathrm{S}_{\theta}(G_t,t)$ is score function to estimate the scores of perturbed graphs $\nabla_{G_t}\log p_t(G_t)$ and $p_t(G_t)$ is the marginal distribution under the forward process at time $t$. In practice, two GNNs are utilized as the score function to denoise both node features and graph structures. $\bar{\mathbf{w}}$ is a reverse time standard Wiener process. % \textcolor{red}{[KX: Actually this section is too abstractive to be understood. 2) if the output of $g_{t}$ is the product of two graphs, how it can be treated as noise to be added over the target graph. 3) what is the definition of $g(t)$? the same with $g_t$?]}

\begin{figure}[!t]    
 \vspace{15pt}
					\centering            % 前面说过，图片放置在中间
					
     % [experiment performance on DrugOOD-IC50-Scaffold]   % 第一张子图的下标（注意：注释要写在[]中括号内）
					
						\label{fig2:a}\includegraphics[scale=0.280]{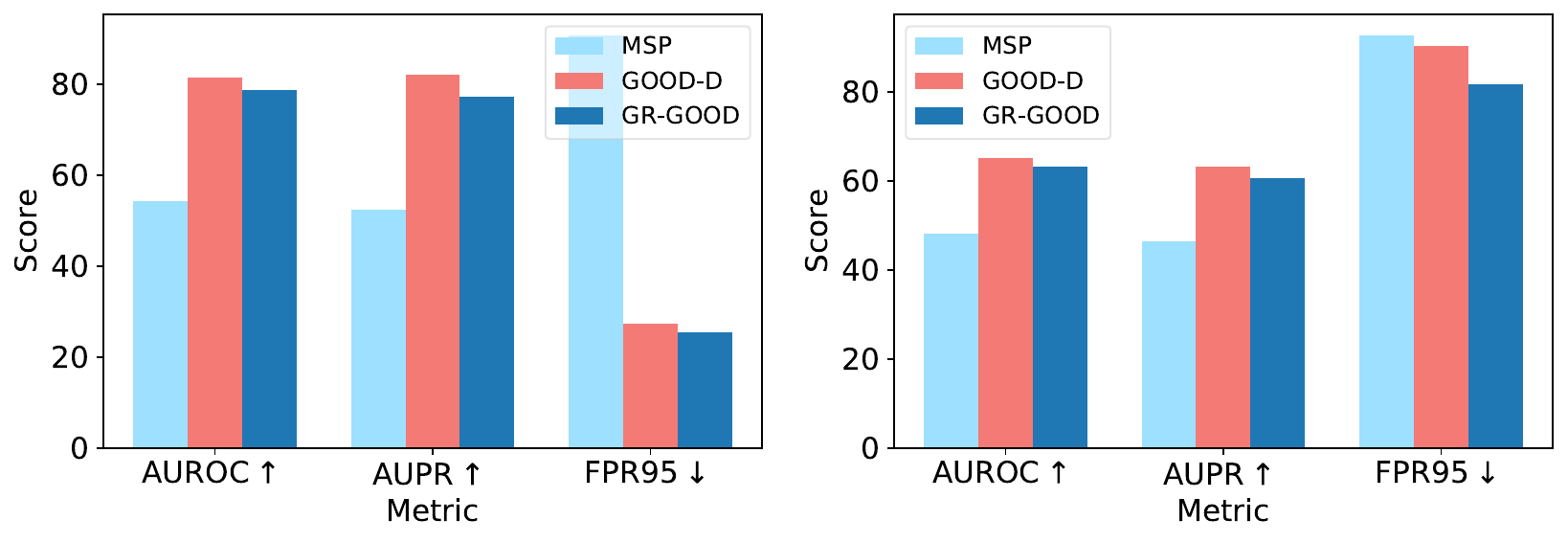}
						
\vspace{-6pt}					
					
					\caption{Validation experiments performed in DrugOOD-IC50-Scaffold (left) and DrugOOD-EC50-Assay (right).} %\textcolor{red}{[KX: color and dont control subfigure height.]}}
     % MSP is the baseline method which utilizes the max softmax value as the OOD score and SOTA is the state-of-the-art graph OOD detection method. Our method outperforms the baseline and performs competitively with SOTA. }    % 整个图片的说明，注释写在{}内
					\label{fig:ob3}           
     \vspace{-9pt}
\end{figure}

\section{Reconstruction of Prototypical Graph for ood detection}
In this section, we first propose a naive graph reconstruction method, termed as GR-MOOD, to analyze its potential and limitations for molecular graph OOD detection.
Then, we propose a novel approach of PGR-MOOD to reconstruct the prototypical graphs of ID samples for effective and efficient OOD detection.
% that does not need to reconstruct any test sample.

\subsection{GR-MOOD}
% We aim to build a naive solution to verify whether reconstruction method is effective.
Inspired by the generative methods \cite{LMD,DiffGuard}, we design a vanilla graph reconstruction model (GR-MOOD) for molecular graph OOD detection. GR-MOOD is pre-trained on a large-scale compound dataset (e.g., QM9 or ZINC) and fine-tuned on $D_{\mathrm{train}}$. 
Considering input graph $G \in D_{\mathrm{test}}$, 
% \textcolor{red}{[KX: Just use $G$ and $\hat{G}$ to represent input and reconstructed graphs]} 
we utilize GR-MOOD to perturb and reconstruct it via:
\begin{align}
    G_{\mathrm{o}} = \operatorname{diffuse}(G,\theta,T),\\
    \hat{G} = \operatorname{denoise}(G_{\mathrm{o}},\theta,T),
\end{align}
where $\theta$ is the parameters of GR-MOOD, and $T$ is the iteration numbers. Function $\operatorname{diffuse}(\cdot)$ 
% $G_{\mathrm{n}}$ is the perturbed graph, $\hat{G}$ is the reconstructed graph,
% \kx{Function, we can omit 'the', the same in the following case.} 
applies Eq.~\eqref{addnoise} to introduce perturbations that transform $G$ into a noised state $G_{\mathrm{o}}$, while function $\operatorname{denoise}(\cdot)$ utilizes Eq.~\eqref{reverse} to reverse the process, effectively denoising $G_{\mathrm{o}}$ to generate reconstruction graph $\hat{G}$.
% reconstruct its original structure $\hat{G}$.

% As we obtain the reconstruction graph $\hat{G}$, the key step is to measure the similarity between $G$ and $\hat{G}$. %Since graphs have not only node features but also the adjacency matrix which represents the topological relationships between nodes, it is challenging to directly compare the similarity between two graphs. Therefore, 
% As we obtain the reconstruction graph $\hat{G}$, 

% The encoding process for a graph $G$ can be written as \textcolor{red}{[KX: Not necessary to have the following function again.]}:
% \begin{align}
%     z_{G} &=  \operatorname{Pooling}\left(\left\{\operatorname{MP}\{X_{v} \}\mid v\in G\right\}\right) ,
% \end{align}
% where $\operatorname{MP}$ and $\operatorname{Pooling}$ are defined the same as in Eq.~\eqref{message-passing} and Eq.~\eqref{pooling}, respectively. 
% The representations of $G_{input}$ and $G_{recon}$ are $z_{input}$ and $z_{recon}$. 
% Upon acquiring the reconstruction graph $\hat{G}$, 
% the key step is to measure the similarity between $G$ and $\hat{G}$.
Upon acquiring the reconstruction graph $\hat{G}$, we utilize a GNN well-trained on the ID dataset to encode both the feature and structure information of  $G$ and $\hat{G}$, whose representations are denoted as $z$ and $\hat{z}$, respectively. 
The cosine similarity between them is treated as OOD judge score and is defined in Eq.~\eqref{judgesim}:
% \kx{Is it the judge score defined in Eq. (1)}:
\begin{align}\label{simgg}
    \operatorname{sim}(G, \hat{G}) = \frac{z \cdot \hat{z}}{\left \| z \right \|\times \left \| \hat{z} \right \|}.
\end{align}
% The value of $\operatorname{sim}(\cdot)$ is treated as the OOD judge score like Eq.~\eqref{judgesim}.

\begin{figure}[t!]    
					\centering            % 前面说过，图片放置在中间
					\subfloat[Performance and Time change with the iteration]   % 第一张子图的下标（注意：注释要写在[]中括号内）
					{
						\label{fig2:a}\includegraphics[scale=0.25]{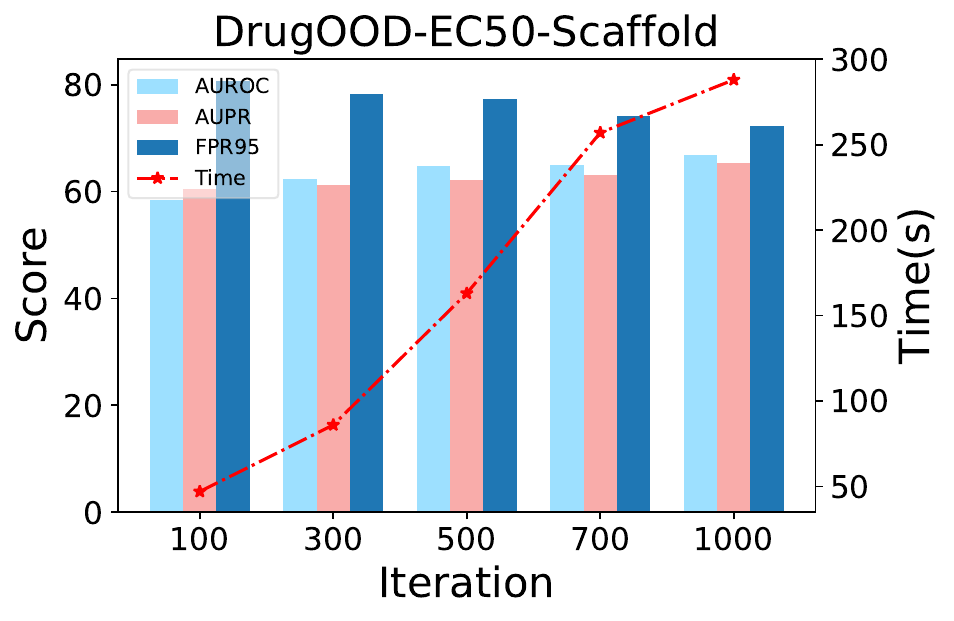}
						
					}\hspace{1mm}
					\subfloat[Reconstruction score distribution for ID and OOD]
					{
						\label{fig2:b}\includegraphics[scale=0.25]{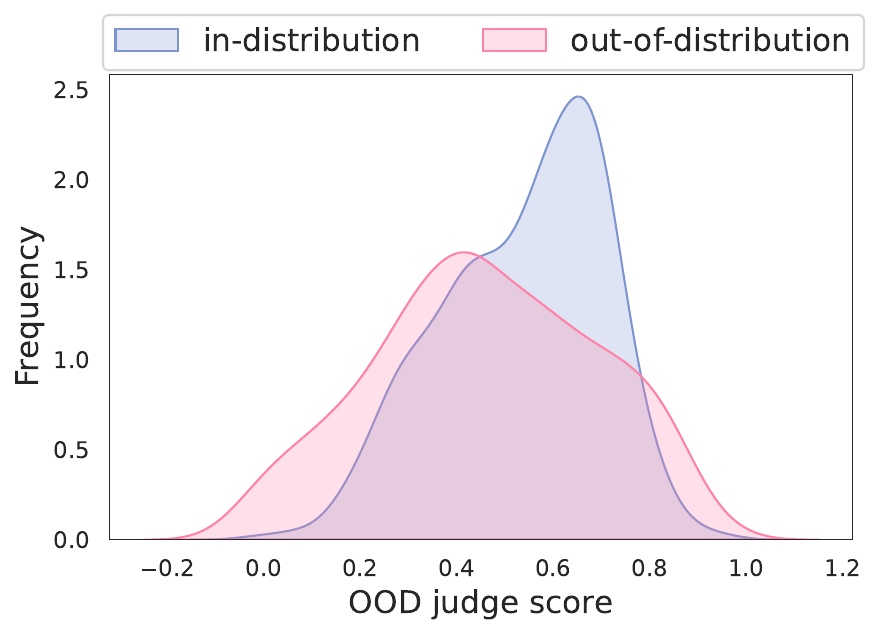}
					}
					
					\vspace{-6pt}
					\caption{Experiments on DrugOOD. (a) Diffusion model requires a large number of iterations to obtain an effective reconstruction. (b) The reconstruction does not yield the discriminative results as expected.}    % 整个图片的说明，注释写在{}内
					\label{fig:ob2}           
     \vspace{-9pt}
\end{figure}
To validate GR-MOOD effectiveness, we conduct experiments on two DrugOOD datasets~\cite{drugood}.
% and we verify its effectiveness by conducting experiments on two DrugOOD datasets~\cite{drugood}.
As shown in Fig.~\ref{fig:ob3}, the performance of GR-MOOD is comparable (e.g., AUROC and AUPR) or even outperforming (e.g., the smaller score of FPR95 is better) than the SOTA method of GOOD-D~\cite{Good-d}.
% for graph OOD detection. 
% We attribute its success to the fact that the diffusion model can effectively learn the mapping between ID distribution and Gaussian distribution (i.e. the distribution of the noise). When performing the $\operatorname{diffuse}(\cdot)$ on the graph, it is equivalent to mapping the graph to a Gaussian distribution, and the $\operatorname{denoise}(\cdot)$ process maps it back to the original distribution, but for OOD, it is unable to map it back to the right position in the distribution where OOD graphs lying in. Therefore, the reconstruction quality of ID graphs is higher than OOD graphs. Although GR-MOOD demonstrated the feasibility of the reconstruction method in molecular graph OOD detection, we found there are still some issues that restrict its performance and lead to high time complexity. 
% The effectiveness of GR-MOOD is ascribed to the diffusion model's proficiency in discerning the correlation between the ID distribution and the Gaussian distribution (i.e., the distribution of the noise). 
The underlying principle is that since GR-MOOD is trained to  reconstruct graphs that align with the ID distribution, OOD samples, due to their inherent dissimilarity from the ID distribution, will typically undergo poorer reconstruction when  being processed. Such discrepancy is quantified as a lower judge score, which signals the presence of an OOD sample.
This mechanism highlights the critical role of diffusion model based reconstruction method in identifying graphs that do not conform to the expected distribution, thereby providing a quantitative basis for distinguishing between ID and OOD samples.

% The function $\operatorname{diffuse}(\cdot)$ effectively transposes the graph onto a Gaussian distribution space, while the $\operatorname{denoise}(\cdot)$ function attempts to revert it to the ID distribution learned by GR-MOOD.  Theoretically, an OOD sample is reconstructed by GR-MOOD to generate a graph that approximates the ID distribution. With the computation of Eq.~\eqref{simgg}, the OOD sample can be easily detected and thus eliminated. 
% \kx{It is still not intuitive why we can detect OOD, should explain the low reconstruction score.}
% However, despite its innovative approach to reconstructing molecular graphs for OOD detection, our evaluation identifies significant limitations in GR-MOOD's performance, including a high time complexity. Against all expectations, GR-MOOD did not achieve a significant breakthrough in detecting molecular graphs. \kx{Move the last few sentences about limitations to the next paragraph.}

% While GR-MOOD's innovative method for reconstructing molecular graphs shows promise for OOD detection, 
\noindent\textbf{Limitation of GR-MOOD:} Despite the intuitive promise of GR-MOOD, our evaluation reveals the non-negligible
limitations in terms of its time efficiency and reconstruction quality measurement. % particularly concerning its high time complexity. 
% Against all expectations, GR-MOOD did not achieve a significant breakthrough in detecting molecular graphs. 
\underline{First}, the primary constraint of GR-MOOD is due to the inherent structural complexity of molecular graphs. As illustrated in Fig.~\ref{fig2:a}, this complexity requires the diffusion model to take an extensive amount of denoising steps to fulfill the reconstruction, improving model performance at the expense of efficiency. Even worse, repeating the generation process for each molecules makes it challenging to scale in the testing phase, which has to screen on a large pool of molecule candidates. \underline{Second}, another issue pertains to the adequacy of the similarity function employed in our model. As depicted in Fig.~\ref{fig2:b}, the reconstruction similarity distributions of ID and OOD samples calculated based on Eq.~\eqref{simgg} % and their respective reconstruction graphs 
are not significantly different~\footnote{There are similar sub-structures among the molecular graphs (e.g., functional groups like benzene rings), resulting in close representations of the OOD and ID samples.}. Since graphs embody as non-Euclidean data, the standard metrics such as cosine similarity impedes the ability to accurately capture the nuances of molecular structure and node features among the molecules. This limitation can result in the consequential loss of detection accuracy.

 \begin{figure*}[ht!]    % 常规操作\begin{figure}开头说明插入图片
					% 后面跟着的[htbp]是图片在文档中放置的位置，也称为浮动体的位置，关于这个我们后面的文章会聊聊，现在不管，照写就是了
					\centering    
% 前面说过，图片放置在中间
	\label{fig:subfig5}\includegraphics[scale=0.39]{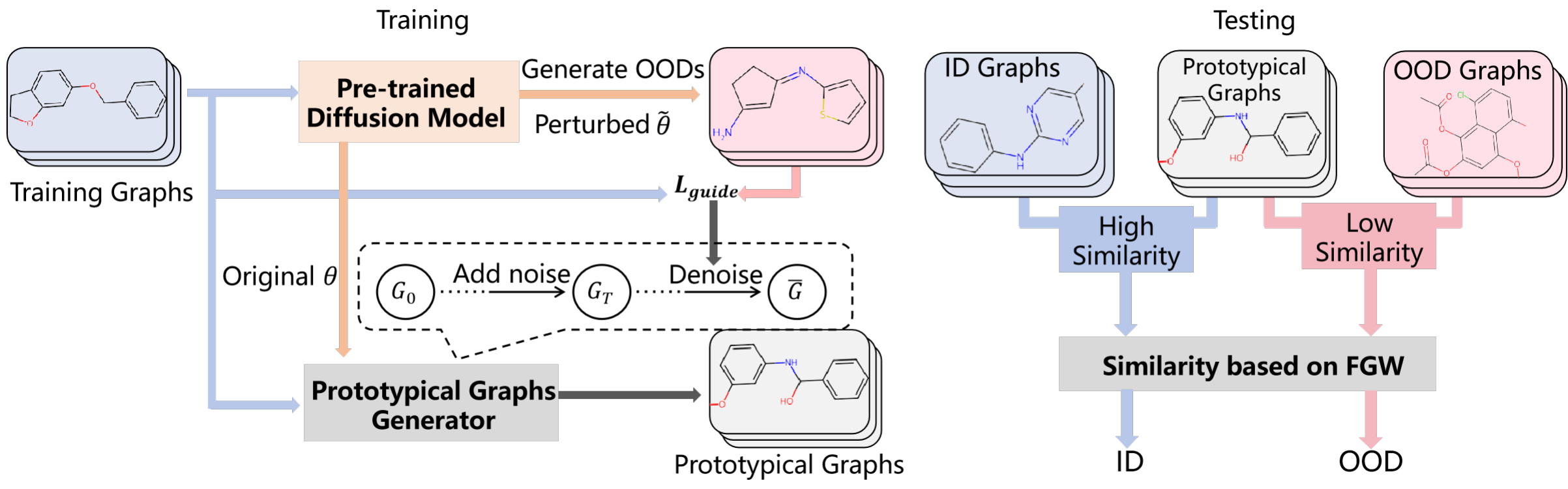}

					\caption{Overview of the proposed PGR-MOOD method. 
     In the training phase, we utilize a pre-trained diffusion model to generate OODs, then calculate $\mathcal{L}_{\mathrm{guide}}$ with OODs and training graphs. 
     Under the guide of $\mathcal{L}_{\mathrm{guide}}$, the prototypical graphs generator generates prototypical graphs $\overline{G}$ as the reconstruction of testing inputs.
     % and such graphs can be treated as the prototype of the reconstruction of testing graphs. 
     In the testing phase, we utilize $\overline{G}$ to calculate the similarity between testing graphs as the OOD judge score.}    % 整个图片的说明，注释写在{}内
					\label{framework}            % 整个图片的标签编号，注意这里跟子图是一样的道理，标签不能重复 
				\end{figure*}
\subsection{PGR-MOOD}
% In the following,
To address the limitations of GR-MOOD, we propose a novel approach based upon diffusion model, PGR-MOOD (Prototypical Graph Reconstruction for Molecular OOD Detection). The innovation of PGR-MOOD has three aspects: A strong similarity function, a prototypical graphs generator, and an efficient and scalable OOD detector. The architecture of PGR-MOOD is shown in Fig.~\ref{framework}.
\\

\noindent\textbf{A Strong Similarity Function based on FGW.} %As a measurement of Euclidean data,
The cosine similarity metric is oriented towards quantifying the angular divergence between two vectors, % positioning it as a tool better suited for the analysis of directional similarity rather than evaluating vector magnitudes. 
while it is not suitable for non-Euclidean data such as graphs.
In fact, measuring the similarity between graphs is equivalent to calculating the their matching degree, the higher the matching degree, the more similar they are. Fused Gromov-Wasserstein ($\operatorname{FGW}$) distance has been proved particularly advantageous for the measurement between graphs. % excels in encapsulating discrepancies not only in structural configurations but also in feature attributes of graphs~\cite{FGW}.
%a metric that exploits the information of the topological structure and the node feature.
% Contrary to the cosine similarity metric, which predominantly quantifies the angular disparity between two vectors. thereby rendering it more appropriate for the assessment of directional congruence as opposed to scalar magnitude, the Fused Gromov-Wasserstein (FGW) distance excels in encapsulating discrepancies not only in structural configurations but also in feature attributes of graphs~\cite{FGW}. \kx{Super long sequence.}
It achieves a balance between the optimal transport (OT) distance with a cost on node features and the Gromov-Wasserstein (GW) distance among the toplogical structures. %It has been proved particularly advantageous for the measurement between graphs. %wherein the interplay between topology and node/edge attributes is of paramount importance.
% Unlike cosine similarity, which primarily measures the angle between two vectors and is thus more suited for comparing overall directionality rather than magnitude, Fused Gromov-Wasserstein distance (FGW) is adept at capturing both structural differences and feature discrepancies in graphs~\cite{FGW}. 
% FGW distance combines Gromov-Wasserstein distance, which measures structural (topological) discrepancies, with traditional optimal transport, which aligns features. 
% This dual approach is particularly beneficial for graph data where both topology and node/edge attributes are crucial. 

Specifically, $\operatorname{FGW}$ 
%allows the graph to be described by a fully supported probability measure over the product space of feature and structure. 
treats the graph associated with topology and node feature as a probability distribution. It allows for the computation of costs between two distributions with optimal coupling, serving as a distance measure between graphs.
% We can take the comparison of both node features and topological structure into the optimal transmission (OT) problem, and the cost calculated between the two distributions with the optimal coupling is used as the distance between the two graphs. 
For two graphs represented in OT format, $G_1 = (A_1,X_1,\mu_{1})$ and $G_2 = ({{A_2}},{{X_2},\mu_{2}})$, their $\operatorname{FGW}$ distance is defined as: 
\begin{align}\label{FGW} \operatorname{FGW}_\alpha(G_1,G_2)&=\min_{\Pi(\mu_1,\mu_2)}\sum_{ijkl}(\alpha(A_1(i,j)-{A_2(k,l)})^2\\ \nonumber
&+(1-\alpha)\|{X}_1(i)-{X}_2(k)\|_2^2)\pi_{ik}\pi_{jl},
\end{align}
where $A_1(i,j)$ represents the element of the $i$-th row and $j$-th column in $A_1$, $X_1(i)$ represents the $i$th row vector of $X$,  $\alpha \in [0,1]$ is a parameter to balance the structure term and the feature term, $\Pi(\mu_1,\mu_2)=\{\pi\in R_{+}^{m\times n} \mathrm{s.t.}, \sum_{i=1}^{m}\pi_{i,j}=\mu_2(j),\sum_{j=1}^{n}\pi_{i,j}=\mu_1(i)\}$ is the set of all admissible couplings between $\mu_1$ and ${\mu_2}$.
$\operatorname{FGW}(\cdot)$ metric exhibits optimal performance in directly discerning both structural variances and feature disparities between graphs. 
% A lower $\operatorname{FGW}(\cdot)$ value between two graphs indicates a greater similarity, allowing for the alignment of the input graph $G$ with its reconstruction $\hat{G}$ without any loss of information.
% A diminutive value of $\operatorname{FGW}(\cdot)$ between two graphs signifies a heightened degree of similarity, thereby enabling its application in aligning the original graph with its reconstructed counterpart without the forfeiture of information. 
% Given that the $\operatorname{FGW}$ metric inherently functions as a measure of dissimilarity, inversely proportional to the anticipated similarity, the employment of its reciprocal facilitates a more intuitive quantification of genuine similarity.
\\
% $\operatorname{FGW}(\cdot)$ is flexible to capture structural and attribute differences between graphs directly and can also be computed for two graphs with different numbers of nodes. 
% The smaller the $\operatorname{FGW}(\cdot)$ value between two graphs, the higher their similarity, thus we can use it to compare the original graph with the reconstruction without losing any information. Since it is the distance function which is the opposite of what we expected we can use the inverse of $\operatorname{FGW}$ distance as the real 
% Once we have a proper similarity function, our next target is to solve the issue of the prohibitive time complexity of performing reconstruction with the diffusion model on all test graphs (aka. $G \in D_{\mathrm{test}}$) as we mentioned in GR-MOOD. 

\noindent\textbf{A Prototypical Graphs Generator.} 
% \kx{Add one or two sentences of motivations to generate proto graphs.}
%The naive diffusion model GR-MOOD only learned how to generates a reconstruction graph that belong to the distribution of $D_{\mathrm{train}}$, but not guaranteed to its quality.
The naive diffusion model of GR-MOOD reconstructs graph that favors the distribution of the input samples, instead of following the distribution learned during the training phase. It misleads the detector's judgment on the OOD samples.
To address this challenge, we propose a prototypical graphs generator, which generates prototypical graphs satisfying the following two properties:
% The generating of a reconstruction graph using the diffusion model should be highly informative, embodying two essential properties: 
\ding{192}\label{proerty1} \textit{For any input graph $G_{\mathrm{in}}\in D_{\mathrm{in}}$, where $D_{\mathrm{in}}$ represents all ID graphs, the prototypical graph ought to closely resemble the graph $G_{\mathrm{in}}$}.
\ding{193}\label{proerty2} \textit{For any input $G_{\mathrm{out}}\in D_{\mathrm{out}}$, where $D_{\mathrm{out}}$ represents all OOD graphs, the prototypical graph should exhibit significant deviation from the graph $G_{\mathrm{out}}$}. Consequently, the goal is to generate a prototypical graph $\overline{G}$ which is close to the ID graphs and far away from the OOD graphs.

To satisfy Property~\ding{192}, Eq.~\eqref{FGW} is utilized as the distance metric, and the loss function $\mathcal{L}_{\mathrm{ID}}$ is formulated to guide the denosing process at the generator:
\begin{align}\label{LID}
    \mathcal{L}_{\mathrm{ID}} = \mathbb{E}_{G_{\mathrm{in}}\sim D^{\mathrm{in}}_{\mathrm{train}}}[\operatorname{FGW}(G_{\mathrm{in}},\overline{G})].
\end{align}
Similarly to comply with Property~\ding{193}, we introduce loss function $\mathcal{L}_{\mathrm{OOD}}$ to enhance the distance between $\overline{G}$ from OOD samples:
% Eq.~\eqref{LID}, an opposite Eq.~\eqref{FGW} can be employed to measure the distance $\overline{G}$ from OOD samples, thus satisfying Property~\ding{193}:
 \begin{align}\label{LOOD}
    \mathcal{L}_{\mathrm{OOD}} = &-\mathbb{E}_{G_{\mathrm{out}}\sim D^{\mathrm{out}}_\mathrm{{train}}}[\operatorname{FGW}(G_{\mathrm{out}},\overline{G})].
\end{align}

Note that OOD graphs $G_{\mathrm{out}}$ are unreachable during the training phase, precluding the direct formulation of $\mathcal{L}_{\mathrm{OOD}}$. Consequently, it becomes imperative to synthesize graphs as proxies for the absent OOD samples.
Recalling the pre-trained diffusion model $\mathrm{F_M}$ in Eq.~\eqref{reverse}, it adopts socre function $\mathrm{S}_{\theta}$ to generate graph. The parameter weights  of $\mathrm{S}_{\theta}$ is given by $\theta_{\mathrm{M}}=\{\theta_{\mathrm{M}}^{(l)}\}_{l=1}^{L}$, where $\theta_{\mathrm{M}}^{(l)}$ represent the parameters of the $l$-th score function. 
We propose to directly perturb parameters $\theta_{\mathrm{M}}$ for generating OOD graphs $G_{\mathrm{out}}$:
\begin{align}\label{perturbationweights}
    \Tilde{\theta}_{\mathrm{M}} =\{ \theta_{\mathrm{M}}^{(l)}(I+\alpha P^{(l)})\}^{L}_{l=1},
\end{align}
where $\alpha >0$ is perturbation strength, $I$ is identity matrix, and $P^{(l)}$ is perturbation matrix. 
By perturbing the parameters $\theta_\mathrm{M}$, a new score function $\mathrm{S}_{\Tilde{\theta}}(\cdot)$ is derived. 
Experimental observations (w/o $\mathcal{L}_{\mathrm{OOD}}$ of Table~\ref{ablation}) reveal that $\mathrm{S}_{\Tilde{\theta}}(\cdot)$ can induce a deviation in the denoising trajectory away from the original data distribution, thereby enabling the diffusion model to generate $G_{\mathrm{out}}$ during the training phase. 
In light of these researches, a composite loss function $\mathcal{L}_{\mathrm{guide}}$ is formulated by integrating both $\mathcal{L}_{\mathrm{OOD}}$ and $\mathcal{L}_{\mathrm{ID}}$: % serving as the overarching objective:
% We can get a novel score function $\mathrm{S}_{\Tilde{\theta}}$.
% Through experiments, we find that  $\mathrm{S}_{\Tilde{\theta}}$ can make the denoising direction deviate from the data distribution from which the input comes and make the diffusion model generate OOD graphs $G_{\mathrm{out}}$ in the training phase. In consideration of the above, we combine $\mathcal{L}_{\mathrm{OOD}}$ and $\mathcal{L}_{\mathrm{ID}}$ as the final loss $\mathcal{L}_{\mathrm{guide}}$:
 \begin{align}
    \mathcal{L}_{\mathrm{guide}} = \mathcal{L}_{\mathrm{ID}}+\mathcal{L}_{\mathrm{OOD}}.
\end{align}
It is leveraged to guides the training of Prototypical Graphs Generator $\mathrm{F_{PG}}$, which has the same architecture and initial parameters $\theta$ with $\mathrm{F_M}$, to generate prototypical graph $\overline{G}$. The generation of $\overline{G}$ by $\mathrm{F_{PG}}$ unfolds in two phases:
Firstly, in contrast to generating directly from Gaussian noise, a graph $G_0$ from $D_{\mathrm{train}}$ is randomly chosen as the start point of generation. We then add $T$-step noise according to Eq.~\eqref{addnoise} to get the final noise graph $G_{T}$ (i.e., $G_0 \to G_{T}$). 
Secondly, $\mathcal{L}_{\mathrm{guide}}$ guides the denoising step of diffusion model to generate prototype graph $\overline{G}$:
% $\mathrm{F_{PG}}$ generate $\overline{G}$ has two steps: 
% Firstly, in contrast to generating directly from noise distribution,  we randomly select a graph in $D_{\mathrm{train}}$ as the start of the generation, then add noise in $T$ steps according to Eq.~\eqref{addnoise} to get the final noise graph ${G}_{T}$. Secondly, we integrate $\mathcal{L}_{\mathrm{guide}}$ into the denoise step of the diffusion model to generate the $\overline{G}$:
\begin{align}\label{novelreverse}
    \mathrm{d}\boldsymbol{{G}}_{t}=[\mathbf{f}_{t}(\boldsymbol{{G}}_{t})-g(t)^2 (\mathrm{S}_{{\theta}}(\boldsymbol{{G}}_{t},t)- \\ \nonumber \nabla_{{G}_{t}}\mathcal{L}_{\mathrm{guide}}(\boldsymbol{{G}}_{t}))]\mathrm{d}t+g(t)\mathrm{d}\overline{\mathbf{w}},
\end{align}
where $t$ is the indicator of the denoise step and varies from $T$ to $0$. 
The prototype graph $\overline{G}$ generated by the above equation can be viewed as the reconstruction of both ID and OOD graphs, but has better discrimination than the reconstruction generated in GR-MOOD. To further reduce the computation, rather than utilizing the entirety of $D_{\mathrm{train}}$, a fixed batch-size dataset $D_\mathrm{\mathrm{batch}}$ is employed for the computation of $\mathcal{L}_{\mathrm{ID}}$. Each $D_{\mathrm{batch}}$ can generate one $\overline{G}$, and they are combined  to formulate a list $PL=\{ \overline{G}^{(i)}\}^{I}_{i=1}$, $I=\lceil \frac{|D_{\mathrm{train}}|}{|D_{\mathrm{batch}}|} \rceil$. 
%Meanwhile, each $D_{batch}$ can generate one $\overline{G}$, and combine them to be a list $PL=\{ \overline{G}^{(i)}\}^{I}_{i=1}$, $I=\lceil \frac{|D_{\mathrm{train}}|}{|D_{\mathrm{batch}}|} \rceil$. 
% The whole process without any parameters needs to be learned and thus can be used as a pre-processing operation. 
\\
%Specifically, we first randomly select a graph in $D_{\mathrm{train}}$ with median node numbers as the original input $G_{0}$ of the diffusion model, then add noise in $T$ steps according to Eq. \ref{addnoise} to obtain ${G}_{T}$. Secondly we perturb the parameters of diffusion model and generate OOD graphs $G_{o}$. Finally we utilize all graphs from 
%$D_{in}$, $G_{ood}$ and ${G}_{T}$ to generate $\overline{G}$ according to Eq. \ref{novelreverse}.

\noindent\textbf{An Efficient and Scalable OOD Detector.} 
Diffusion models require significant time and memory resources during the testing phases because they need to generate a reconstructed graph  for each input.
% Although the generated prototypical graphs $\overline{G}$ satisfy Property\ding{192}\ding{193} perfectly, the diffusion model still costs a lot of time in the training and testing phases. 
% The diffusion model incurs substantial computational overhead during both the training and testing phases. 
To alleviate this computational burden, PGR-MOOD 
%as we have the prototype graph $\overline{G}$ which obey property~\ding{192} and ~\ding{193}, directly calculate the similarity between test graphs and  $\overline{G}$ as the final judge score can avoid any generating process. rather than utilizing the entirety of $D_{\mathrm{train}}$, a fixed batch-size dataset $D_\mathrm{{batch}}$ is employed for the computation of $\mathcal{L}_{\mathrm{ID}}$. 
%Meanwhile, each $D_{batch}$ can generate one $\overline{G}$, and combine them to be a list $PL=\{ \overline{G}^{(i)}\}^{I}_{i=1}$, $I=\lceil \frac{|D_{\mathrm{train}}|}{|D_{\mathrm{batch}}|} \rceil$. 
eliminates the necessity of graph reconstruction in the testing phase via preparing the prototypical graphs in the training phase. PGR-MOOD leverages the $\overline{G}$ within list $PL$ to conduct the similarity measurement with every new test sample. The maximum similarity is employed as the definitive judge score for OOD detection:
\begin{align}\label{templateOOD}
    J(G) = \max_{\overline{G}\sim PL}[\operatorname{sim}(\overline{G},G_{\mathrm{test}})], \ G_{\mathrm{test}}\in D_{\mathrm{test}}.
\end{align}
where $\operatorname{sim}$ ($\cdot$) is the similarity function based on the inverse of $\operatorname{FGW}$ distance.
\begin{algorithm}[h]
  \caption{PGR-MOOD}
  \begin{algorithmic}[1]
    \Require
      A Pre-trained diffusion models $\mathrm{F_M}$;
      The data loader of in-domain training set $D_{\mathrm{train}}$;
      An empty prototypical graphs lists $PL$;
      Denoise step $T$.
    \Ensure
        Prototypical graphs lists $PL$;
    \State Utilize Eq.~\eqref{perturbationweights} to perturb the parameters of $\mathrm{F_M}$ to get $\Tilde{\theta}$ ;
       \State Generate $G_{\mathrm{OOD}}$ though $\mathrm{F_M}$ with parameters $\Tilde{\theta}$;
    \For{$G_\mathrm{{batch}}$ in $D_{\mathrm{train}}$}
       \State Random select a graph $G_{0}$ from $G_{\mathrm{batch}}$; 
       \State Utilize Eq.~\eqref{addnoise} to calculate noise graph $G_{T}$ with $G_{0}$; 
       \For{t in T to 1}
      \State Compute $\mathcal{L}_{\mathrm{guide}}$ with $G_{\mathrm{batch}}$ and $G_{\mathrm{OOD}}$.
      \State  %Generate $\overline{G}$ though multiple 
      Perform denoise steps in Eq.~\eqref{novelreverse} with $\mathcal{L}_{\mathrm{guide}}$ and $G_{t}$.
       \EndFor
        % \State Generate $\overline{G}$ and add to $TL$;
        \State Add $\overline{G}$ to $PL$;
      
    \EndFor
    
    \label{code:recentEnd}
  \end{algorithmic}
\end{algorithm}

\begin{table*}[]
			\centering
			\normalsize
			
			\caption{OOD detection performance on the DrugOOD dataset. Scaffold, Size, and Assay are the basis for dividing ID and OOD graphs.  The best and runner-up results are highlighted with bold and \underline{underline}, respectively.
				 }
			\setlength\tabcolsep{4.0pt}
			\renewcommand{\arraystretch}{1.0}
			\begin{tabular}{c|ccc|ccc|ccc}
				
				\hline
				 \multicolumn{10}{c}{DrugOOD-IC50} \\ \hline 
    & \multicolumn{3}{c|}{Scafflod} & \multicolumn{3}{c|}{Size} & \multicolumn{3}{c}{Assay}\\ \hline
				OOD Detector & AUROC $\uparrow$      & AUPR $\uparrow$      & FPR95 $\downarrow$    & AUROC $\uparrow$     & AUPR $\uparrow$    & FPR95  $\downarrow$   & AUROC $\uparrow$     & AUPR $\uparrow$    & FPR95 $\downarrow$   \\ \hline
				MSP     &54.57±9.18 &52.43±6.85 &90.76±4.95        & 52.57±9.07 &57.23±3.25 &88.60±4.75     & 58.19±7.23 &56.38±5.75 &89.20±3.05\\ \hline
				GOOD-D    & \underline{85.40±1.23 }& \underline{87.13±2.31 }& {27.40±2.37 }& \underline{91.55±1.10 }& \underline{87.91±3.74}  & \underline{16.95±0.47}  & \underline{81.35±1.74}& \underline{79.05±0.79}  & \underline{75.02±0.57} \\
				GraphDE   & 69.15±1.11 &67.40±0.51 &80.30±0.33 & 78.72±1.78 &79.36±1.24 &78.97±0.75  & 68.56±1.08 &66.56±0.31 &82.20±0.93\\
				AAGOD    & 84.23±2.97  & 83.96±1.34  & \underline{21.56±1.08} & 84.75±1.23  & 83.32±1.61  & 19.80±0.93  & 71.94±1.45  & 72.86±1.84  & 85.62±2.71 \\ \hline
                OCGIN  & 68.39±4.77 &66.05±5.11 &82.80±7.50 & 70.94±5.09 &68.99±3.72 &74.80±6.46  & 67.53±4.61 &66.95±5.23 &79.80±4.60\\
				GLocalKD   & 63.42±0.60 &58.03±0.64 &70.28±1.83 & 69.44±0.58 &67.29±0.77 &81.13±1.46  & 62.08±0.76 &61.93±0.61 &82.70±1.98 \\ \hline
               GR-MOOD   & 78.82±2.31 &77.35±1.94 &{25.43±1.72} & 68.51±2.65 &69.19±3.01 &70.78±2.33  & 61.91±1.87 &62.95±1.54 &84.87±1.39 \\ \hline
				PGR-MOOD     & \textbf{91.57±1.32} & \textbf{90.12±0.71} & \textbf{19.42±0.22} & \textbf{93.84±1.53} & \textbf{94.85±2.03}  & \textbf{15.57±1.03}  & \textbf{83.72±2.51} & \textbf{80.31±1.44}  & \textbf{64.65±0.57} \\ Improve    & +7.22\% & +3.43\% & -9.89\% & +2.50\%  & +7.08\% & -8.41\%  & +2.91\%   & +1.52\% &-13.80\%  \\  \hline
			\end{tabular}
			
			\begin{tabular}{c|ccc|ccc|ccc}

     \multicolumn{10}{c}{DrugOOD-EC50} \\ \hline & \multicolumn{3}{c|}{Scafflod} & \multicolumn{3}{c|}{Size} & \multicolumn{3}{c}{Assay}\\ \hline
				OOD Detector & AUROC $\uparrow$      & AUPR $\uparrow$      & FPR95 $\downarrow$     & AUROC $\uparrow$     & AUPR $\uparrow$    & FPR95 $\downarrow$    & AUROC $\uparrow$     & AUPR $\uparrow$    & FPR95 $\downarrow$   \\ \hline
				MSP     &57.26±7.25 &57.08±5.94 &87.26±5.12        & 59.18±8.77 &58.41±4.95 &83.76±5.60        & 48.19±9.18 &46.38±6.85 &89.26±4.95\\ \hline
				GOOD-D    & \underline{82.51±1.31} & \underline{81.98±2.71} & \underline{63.21±2.89} & \underline{92.50±1.32} & \underline{88.37±1.26}  & \underline{19.20±0.51}  & 65.20±1.48& 67.22±1.61  & 92.24±3.56 \\
				GraphDE   & 68.55±1.03 & 66.56±1.90 & 82.20±0.74 & 79.64±1.16 & 77.75±1.48  & 59.25±0.57  & 66.24±1.79 & 66.28±0.98  & 80.29±1.04\\
				AAGOD    &{77.17±5.52}  &{75.32±5.56}    &{72.76±4.95}        & 78.72±6.59 &79.23±6.30 &68.66±5.43       & \underline{74.57±9.18} &\underline{72.43±6.85} &\underline{71.83±4.43}\\ \hline
                OCGIN  & 69.01±3.98 &67.83±4.87 &74.79±7.50 & 78.45±5.17 &74.30±3.96 &81.53±5.64  & 71.33±2.85 &70.94±3.69 &80.93±3.55\\
                
				GLocalKD   & 66.59±0.71 &68.64±0.45 &71.22±1.01 & 
                69.59±0.98 &68.72±0.83 &68.70±1.36  & {73.32±1.65} &69.23±1.57 &75.39±2.19\\ \hline
                GR-MOOD  & 71.15±2.50 &73.02±3.21 &81.79±3.58 & 73.80±2.95 &78.49±1.63 &70.96±1.82  & 60.17±1.56 &61.69±10.27 &79.09±1.33 \\ \hline
				PGR-MOOD    & \textbf{87.53±1.31} & \textbf{86.16±0.72} & \textbf{62.82±2.21} & \textbf{97.67±1.54} & \textbf{96.32±1.47}  & \textbf{13.79±1.23}  & \textbf{86.73±3.34} & \textbf{83.56±3.28}  & \textbf{63.74±2.59}\\ Improve & +6.02\%  &+5.09\% & -3.70\% & +5.58\%  & +8.41\%  & -28.10\%  & +16.30\%  & +15.36\%   & +11.22\% \\ \hline
			\end{tabular}
			\label{table_Drugood}
		\end{table*}

  \begin{table*}[]
			\centering
			\normalsize
			
			\caption{Ablation experiment results on four datasets.  }
			\setlength\tabcolsep{5.0pt}
			\renewcommand{\arraystretch}{0.8}

			\begin{tabular}{c|ccc|ccc|ccc}
				
				\hline
				
     \hline & \multicolumn{3}{c|}{AUROC $\uparrow$ } & \multicolumn{3}{c|}{AUPR $\uparrow$ } & \multicolumn{3}{c}{FPR95 $\downarrow$ }\\ \hline
				Dataset & w/o $\mathcal{L}_{\mathrm{ID}}$    & w/o $\mathcal{L}_{\mathrm{OOD}}$ & w/o $\operatorname{FGW}$    & w/o $\mathcal{L}_{\mathrm{ID}}$    & w/o $\mathcal{L}_{\mathrm{OOD}}$ & w/o $\operatorname{FGW}$    & w/o $\mathcal{L}_{\mathrm{ID}}$    & w/o $\mathcal{L}_{\mathrm{OOD}}$ & w/o $\operatorname{FGW}$  \\ \hline
				DrugOOD-EC50     &-4.57 &-2.43 &-0.76        & -7.72 &-2.32 &-4.75        & +5.74 &+2.22 &+1.63\\ 
				DrugOOD-IC50     &-5.14 &-1.75 &-1.24        & -4.26 &-1.98 &-3.62        & +6.83 &+1.77 &+2.36 \\
				GOOD-HIV  &-3.26 &-2.58 &-0.54        & -5.83 &-2.43 &-3.18        & +4.72 &+2.03 &+2.61\\
				GOOD-PCBA   &-5.89 &-1.08 &-2.07        & -6.44 &-3.70 &-4.81        & +3.62 &+1.12 &+2.14\\ \hline
               
			\end{tabular}
			\label{ablation}
		\end{table*}
\vspace{-10pt}
\section{experiment}
% \usepackage{graphicx}
In this section, we verify the effectiveness of PGR-MOOD and GR-MOOD by performing experiments on two graph OOD benchmarks. 
% \begin{itemize}
%     \item Can PGR-MOOD overcome the shortcomings of naive method and achieve better performance.
%     \item Can PGR-MOOD exceed the performance of the previous state of the art methods.
%     \item How do the composed modules impact PGR-MOOD’s performance?
%     \item How do the hyper-parameters impact PGR-MOOD’s performance? 
%     \item Can our PGR-MOOD be more efficient in terms of execution time and memory allocation.
% \end{itemize}
\subsection{Experiment Setup}
% \usepackage{multirow}
% \usepackage{graphicx}

\subsubsection{Datasets} 
%With the increasing attention of researchers on OOD detection in molecular graphs, two benchmarks focused on this task have been proposed: DrugOOD~\cite{drugood} and GOOD~\cite{good}. 
With the increasing attention on OOD detection in the molecular graphs, two benchmarks are proposed, GOOD~\cite{good} and DrugOOD~\cite{drugood}, respectively. These two benchmarks provide the detailed rules to distinguish between ID and OOD. GOOD is built based on the scaffold and size of the molecular graph, and DrugOOD adds an assay on the basis of these two distribution shifts. 
% which indicates that the collected molecular graph features are shifted by different experimental environments. 
We take six datasets from DrugOOD and four datasets from GOOD as our experimental datasets. Please see Appendix~\ref{A.1} for details.

\subsubsection {Baselines Methods}
To verify the performance of our methods, namely GR-MOOD and PGR-MOOD, we use the GNNs’ Max Softmax Score (MSP)~\cite{msp} as a vanilla baseline and then compare with three SOTA graph OOD detection methods (GOOD-D ~\cite{Good-d}, AAGOD~\cite{AAGOD}, and GraphDE~\cite{graphde}). Meanwhile, two graph anomaly detection methods, namely OCGIN~\cite{OCGIN} and GLocalKD~\cite{GLocalKD}, are introduced as the baseline. In addition, as the first molecular graph OOD detection method based on the diffusion model, we also compare the PGR-MOOD with the naive solution GR-MOOD to verify whether its limitations have been solved. Please see Appendix~\ref{A.2} for details.
%To further verify the performance of our method, we use the GNNs’ max softmax score (MSP)~\cite{msp} as a vanilla baseline and then compare with three SOTA graph OOD detection methods (GOOD-D, AAGOD, GraphDE~\cite{Good-d,AAGOD,graphde}). Also, two graph anomaly detection methods OCGIN~\cite{OCGIN}, and GLocalKD~\cite{GLocalKD} are introduced as the baseline.
% which need to modify the GNN's architecture or adjusting the training strategy. 
% Two graph anomaly detection methods (OCGIN, GLocalKD) also be considered to be the baseline since OOD graphs can also be treated as the outlier.

\subsubsection{Implementation Details}
For our methods, we utilize the diffusion model GDSS~\cite{gdss} as the backbone
which achieves stat-of-the-art performance on graph generation. GDSS is pre-trained on the QM9 dataset, which comprises a large collection of organic molecules with 113k samples. Following the setting of GraphDE, we perform 10 random trials and report the average accuracy on the test set, along with $95\%$ confidence intervals. During training, we set $\alpha$ to $0.5$ to balance the topological structure and node features  when computing the FGW distance. We set $D_{batch}$ to 128 and the number of perturbation steps $T\in [1,10]$ to reduce memory allocation and computation complexity. For all baseline methods, we follow settings reported in their papers. All the experiments are implemented by PyTorch, and run on an NVIDIA TITAN-RTX (24G) GPU.

\subsection{Performance Analysis}

\textbf{Q: Whether PGR-MOOD achieves the best performance on the OOD detection in molecular graphs?} Yes, we utilize the new loss function $\mathcal{L}_{\mathrm{guide}}$ to guide the diffusion model to generate prototypical graphs that are more representative of all ID samples, and more easily detect OOD samples. 

% \subsubsection{Comparison with the naive solution}
% From the results, we demonstrate that our method has a very obvious improvement compared with a naive solution on all ten datasets. 
\noindent$\rhd$ \textsf{Comparison with the naive solution.}
As shown in Table~\ref{table_Drugood} and Table~\ref{table_good}, compared with GR-MOOD on six datasets of DrugOOD, PGR-MOOD enhances the average AUC and AUPR by $32.76\%$ and $29.54\%$, and reduces the average FPR95 by $45.65\%$. These results demonstrate that the prototypical graphs of PGR-MOOD generated with the FGW similarity function are more suitable for distinguishing the original input graphs in the testing phase.
\noindent$\rhd$ \textsf{Comparison with the State-of-the-art Methods.} To verify the superiority of our method, we compare it with the previous SOTA methods.
As shown in the last row of Table~\ref{table_Drugood} and Table~\ref{table_good}, 
our method achieves SOTA results on all datasets. The average improvements against the previous SOTA are $8.54\%$ of AUC and $8.15\%$ of AUPR, and the average reduction on FPR95 is $13.7\%$.
% For six datasets of DrugOOD, we outperform the SOTA methods on five datasets and enhance the average AUC and AUPR by $6.76\%$ and $6.68\%$, reducing the average FPR95 by $12.20\%$ relatively. For four datasets of GOOD, we achieved out-performance on all datasets and the relative improvements against the previous SOTA are $11.25\%$ and $10.14\%$ on AUC and AUPR, the average reduction on FPR95 is $15.30\%$.
%\textcolor{red}{We attribute this to the fact that the diffusion model can generate valid prototypical graphs under the guidance of $\mathcal{L}_{\mathrm{guide}}$ and $\operatorname{FGW}$ can accurately measure the similarity between it and the original input graph.}
We attribute these results to the fact that the prototypical graphs generated by PGR-MOOD can enlarge the judge score gap between ID and OOD which satisfies the requirement of optimal OOD detector.

\begin{table*}[ht!]
    \centering
    \caption{OOD detection performance on the GOOD dataset. Scaffold and Size are the basis for dividing ID and OOD graphs. Best and runner-up results are highlighted with bold and \underline{underline}, respectively.}
    \setlength\tabcolsep{4.0pt} % Adjust table padding as necessary
    \renewcommand{\arraystretch}{1.0} % Adjust line spacing as necessary
    \begin{tabular}{c|c|c|c|c|c|c|c|c|c|c}
    \hline
    \multicolumn{11}{c}{GOOD-HIV}                     \\ \hline
    Dataset & {Metric} 
    & MSP & GOOD-D & GraphDE & AAGOD & OCGIN & GLocalKD & GR-MOOD&PGR-MOOD &Improve\\ \hline
    \multirow{3}{*}{Scaffold} & AUROC $\uparrow$  & 58.55±9.18 & 62.42±1.89 & 65.66±1.69 & \underline{74.81±1.56} & 66.29±4.35 & 64.76±0.34 &61.22±2.68& \textbf{85.57±1.32}&+14.38\% \\ 
    & AUPR$\uparrow$  & 58.34±6.85 & 69.60±2.03 & 60.94±0.48 & \underline{72.51±1.99} & 65.45±5.98 & 65.92±0.64 & 60.53±1.94&\textbf{85.12±0.71}&+12.61\% \\
    & FPR95$\downarrow$ & 93.40±4.95 & 87.75±0.35 & 88.40±0.43 & \underline{76.71±1.82} & 85.65±6.74 & 83.98±0.89 & 87.35±1.66&\textbf{66.50±2.01}&-13.31\% \\ \hline
    \multirow{3}{*}{Size} & AUROC$\uparrow$  & 54.96±9.07 & \underline{72.23±1.54} & 66.72±1.13 & 63.44±1.92 & 65.04±4.65 & 68.49±1.22 & 69.67±2.71&\textbf{88.43±2.37}&+22.47\% \\
    & AUPR$\uparrow$  & 54.09±3.25 & \underline{76.12±1.26} & 65.55±0.30 & 60.02±1.88 & 64.67±4.03 & 68.23±0.97 &71.76±2.39& \textbf{87.77±2.18}&+15.30\% \\
    & FPR95$\downarrow$ & 97.80±4.75 & \underline{68.74±3.25} & 72.20±0.89 & 75.97±1.15 & 73.64±5.86 & 76.13±1.55 &60.56±2.91& \textbf{65.17±2.21}&-5.17\% \\
    \hline
\end{tabular}

\begin{tabular}{c|c|c|c|c|c|c|c|c|c|c}
        
        \multicolumn{11}{c}{GOOD-PCBA}                     \\ \hline
    Dataset & {Metric} 
    & MSP & GOOD-D & GraphDE & AAGOD & OCGIN & GLocalKD & GR-MOOD&PGR-MOOD &Improve\\ \hline
    \multirow{3}{*}{Scaffold} &
        AUROC $\uparrow$  & 54.57±9.07 & \underline{85.69±1.16} & 68.45±1.23 & 79.06±0.48 & 69.50±3.17 & 70.90±1.68 & 70.07±0.60&\textbf{86.57±1.32}&+1.02\% \\
       &AUPR $\uparrow$  & 52.43±6.21 & \underline{86.97±1.76} & 66.07±0.32 & 72.70±0.30 & 68.34±4.11 & 73.56±1.64 &71.90±0.64& \textbf{88.12±0.71}&+1.32\% \\
        &FPR95 $\downarrow$  & 90.76±4.36 & \underline{16.04±1.90} & 82.34±0.67 & 60.37±0.58 & 87.94±6.98 & 39.57±1.44 & 55.42±1.89&\textbf{15.01±0.32}&-6.04\% \\
        \hline
        \multirow{3}{*}{Size}&AUROC $\uparrow$  & 58.57±8.99 & \underline{78.31±1.19} & 66.24±1.90 & 64.90±1.71 & 70.61±3.25 & 73.58±0.50 &71.49±0.78& \textbf{83.84±1.53} &+7.06\%\\
        &AUPR $\uparrow$  & 57.23±3.25 & \underline{76.21±1.61} & 64.58±0.21 & 67.24±0.87 & 72.21±3.91 & 67.40±0.91 & 75.31±1.09&\textbf{84.85±2.03}&+11.33\% \\
        &FPR95 $\downarrow$  & 88.60±4.75 & \underline{27.30±1.72} & 88.45±0.29 & 60.03±1.06 & 63.80±4.47 & 60.29±0.89 & 46.37±1.29&\textbf{17.01±0.17}&-37.61\% \\ \hline
    \end{tabular}
    \label{table_good}
\end{table*}

\subsection{Visualization of Score Gap}
\begin{figure}[!t]    
					\centering            % 前面说过，图片放置在中间
   
					\subfloat[HIV-Scaffold]   % 第一张子图的下标（注意：注释要写在[]中括号内）
					{
						\includegraphics[width=0.152\textwidth,height=0.11\textheight]{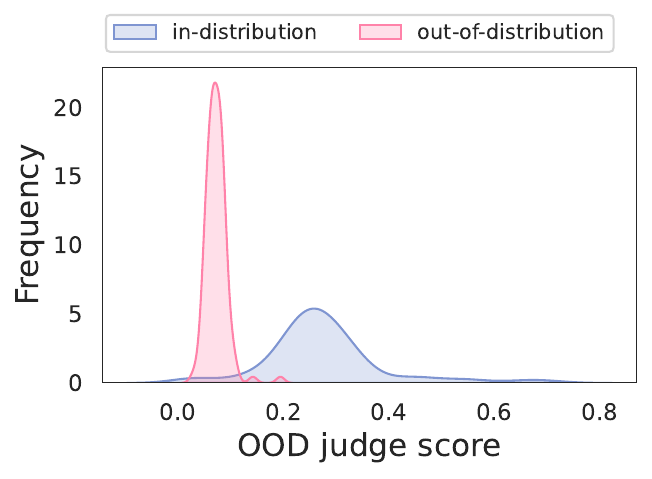}
						
					}
      % \hspace{-10pt}
					\subfloat[IC50-Scaffold]
					{
						\includegraphics[width=0.152\textwidth,height=0.11\textheight]{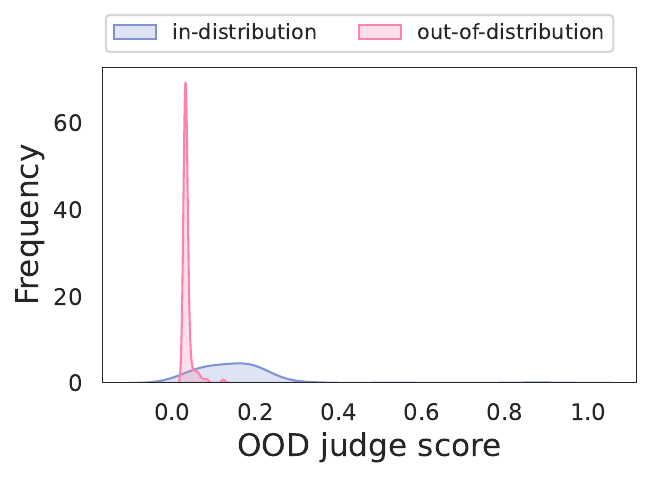}
					}
    % \hspace{-10pt}
					\subfloat[IC50-Size]
					{
						\includegraphics[width=0.152\textwidth,height=0.11\textheight]{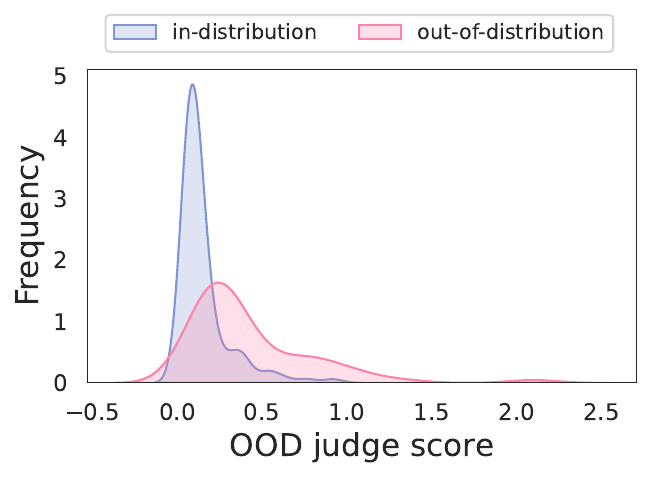}
					}
  \vspace{-10pt}
					\caption{OOD judge score distributions on three datasets. }    % 整个图片的说明，注释写在{}内
					\label{fig:scoregap}  
   \vspace{-10pt}
				\end{figure}

\textbf{Q: Whether PGR-MOOD can enlarge the judge score gap between ID and OOD graphs?} Yes, we calculate the similarity between the prototypical graphs and test graphs, which has a massive difference for ID and OOD. 
A more significant gap between ID and OOD graphs corresponds to a better graph OOD detector. We present the scoring distributions on two datasets in Fig.~\ref{fig:scoregap}. The ID and OOD are perfectly separated into two distinct distributions, so we can use a simple threshold for OOD detection and achieve SOTA performance.

\subsection{Ablation Experiment}
\textbf{Q: Whether each module in PGR-MOOD contribute to effectively discriminating OOD molecular graphs?} Yes, we conduct experiments on four datasets to verify the role of $\mathcal{L}_{\mathrm{ID}}$, $\mathcal{L}_{\mathrm{OOD}}$, and $\operatorname{FGW}$ modules in PRG-MGOD. The results are shown in Table~\ref{ablation}.

% We experiment on four datasets which all use scaffold as the domain partition criterion.

% \subsubsection{Ablation on $\mathcal{L}_{\mathrm{ID}}$ and $\mathcal{L}_{\mathrm{OOD}}$} 
\noindent$\rhd$ \textsf{Ablation on $\mathcal{L}_{\mathrm{ID}}$ and $\mathcal{L}_{\mathrm{OOD}}$.} We remove $\mathcal{L}_{\mathrm{ID}}$ and $\mathcal{L}_{\mathrm{OOD}}$ in the $\mathcal{L}_{\mathrm{guide}}$ respectively to explore their impacts on the performance of OOD detection. We find that merely enlarging the distance between the prototypical graph from  OOD samples (w/o $\mathcal{L}_{\mathrm{ID}}$) or bringing it closer to ID samples (w/o $\mathcal{L}_{\mathrm{OOD}}$) significantly undermines the performance of PGR-MOOD. %In the w/o $\mathcal{L}_{\mathrm{ID}}$ case by its obviousness. 
This fully confirms that the Property\ding{192} and Property\ding{193} are valid and correct. These results demonstrate that the composition of $\mathcal{L}_{\mathrm{ID}}$ and $\mathcal{L}_{\mathrm{OOD}}$ can generate prototypical graphs $\overline{G}$ with different similarity measurement for ID and OOD graphs in the testing phase.

% only moving the prototypical graph away from the OOD (w/o $\mathcal{L}_{\mathrm{ID}}$) or pulling it into the ID (w/o $\mathcal{L}_{\mathrm{OOD}}$) all affect the PGR-MOOD. 
% As shown in Table \ref{ablation}, we can find w/o $\mathcal{L}_{\mathrm{ID}}$ both AUROC and AUPR decrease average $4.7\%$ and $6.2\%$ and FPR95 increase average $5.2\%$ on all datasets. 
% If without $\mathcal{L}_{\mathrm{OOD}}$, both AUROC and AUPR decrease average $1.9\%$ and $2.6\%$, FPR95 increases average $1.8\%$ on all datasets. 
% These results demonstrate that only the composition of $\mathcal{L}_{\mathrm{ID}}$ and $\mathcal{L}_{\mathrm{OOD}}$ can generate prototypical graph $\overline{G}$ with different similarity for ID and OOD graphs.

% \subsubsection{Ablation on $\operatorname{FGW}$}
\noindent$\rhd$ \textsf{Ablation on $\operatorname{FGW}$.} 
We replace the $\operatorname{sim}(\cdot)$ function based on $\operatorname{FGW}$ in Eq.~\eqref{templateOOD} with Eq.~\eqref{simgg} of GR-MOOD to explore its importance on the performance of OOD detection. We find that the $\operatorname{FGW}$ is even more influential than $\mathcal{L}_{\mathrm{OOD}}$ on all datasets with different metrics. These experimental results demonstrate that a proper similarity measurement is necessary and the $\operatorname{FGW}$ can thoroughly evaluate the similarity between two graphs by considering both their structure and features.

\begin{figure}[t!]   
\vspace{-5pt}
					\centering            % 前面说过，图片放置在中间
     % \subfloat[GOOD-HIV-Scaffold]
					   % 第一张子图的下标（注意：注释要写在[]中括号内）
					
					\label{fig:subfig4}\includegraphics[width=0.48\textwidth,height=0.12\textheight]{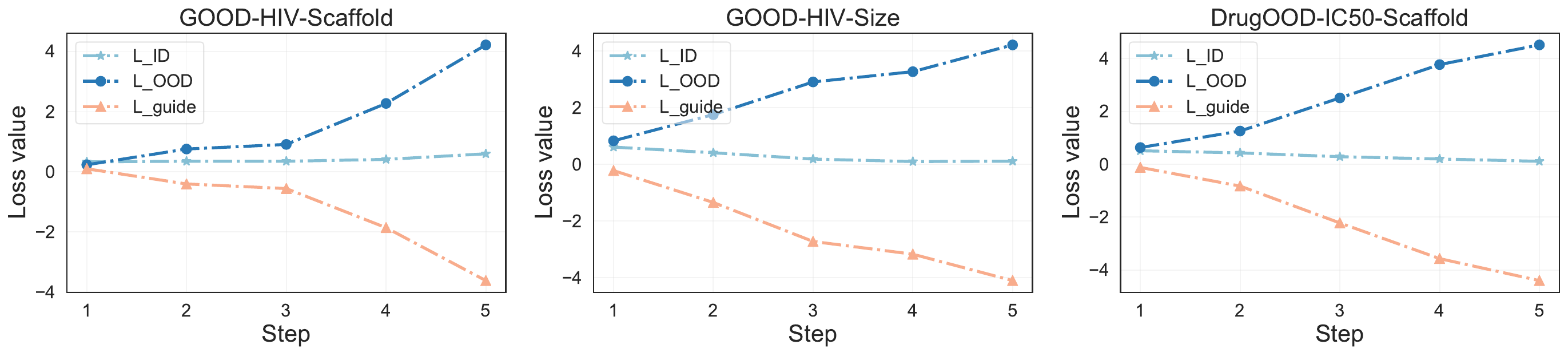}

    \vspace{-13pt}
					\caption{Loss variation during generation on three datasets. }
     % \kx{Font size, and subfigure height.}}    % 整个图片的说明，注释写在{}内
					\label{figablation}           
     \vspace{-20pt}
\end{figure}

% As shown in Table \ref{ablation}, we can find without $\operatorname{FGW}$, both AUROC and AUPR decrease average $1.2\%$ and $4.09$, FPR95 increase average $2.1\%$ on all datasets. These results demonstrate that a proper similarity is necessary and due to $\operatorname{FGW}$ can comprehensively compare the similarity between two graphs through both the structure and features of the graph.

% \subsubsection{Exploration of $\mathcal{L}_{\mathrm{guide}}$} 

\noindent\textbf{Q: Whether the prototypical graphs $\overline{G}$ generated by $\mathcal{L}_{\mathrm{guide}}$-guided PGR-MOOD follow the Properties \ding{192} and \ding{193} ?} Yes, the prototypical graphs $\overline{G}$ effectively reduce the distance with the ID graphs and significantly increase the separation from the OOD graphs. 
To validate the impact of $\mathcal{L}_{\mathrm{guide}}$, its trend is monitored throughout the generation phase, as depicted in Fig.~\ref{figablation}. Here, $\mathcal{L}_{\mathrm{ID}}$ and $\mathcal{L}_{\mathrm{OOD}}$ are computed using Eq.~\eqref{LID} and Eq.~\eqref{LOOD} and they represent the distance between $\overline{G}$ and all graphs belong to ID and OOD, respectively. As the generation progresses, $\mathcal{L}_{\mathrm{ID}}$ steadily decreases towards 0, whereas $\mathcal{L}_{\mathrm{OOD}}$ escalates sharply. This observation aligns seamlessly with the foundational principles of PGR-MOOD.

\begin{figure}[t!]    
					\centering            % 前面说过，图片放置在中间
     \includegraphics[scale=0.21]{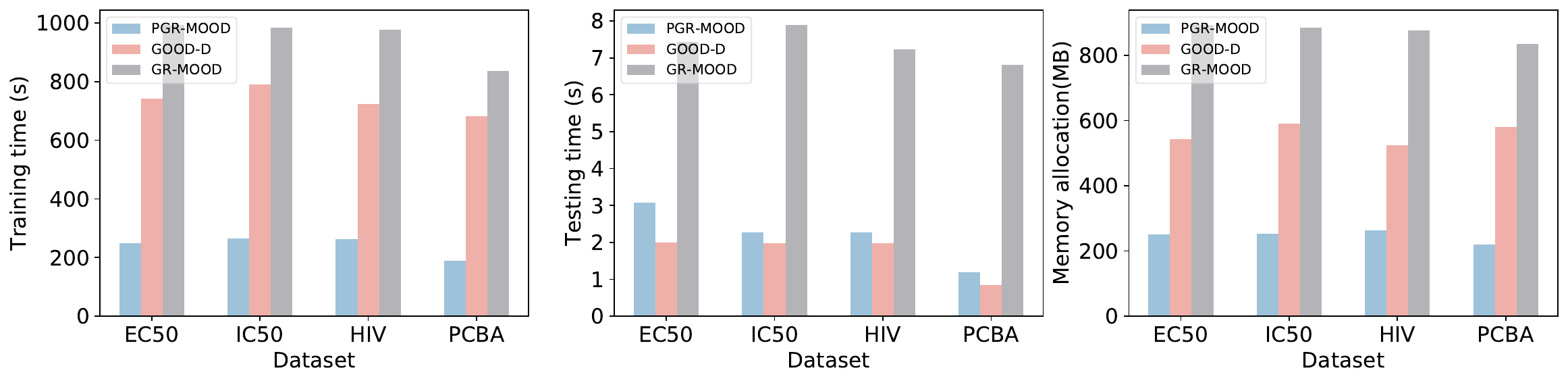}
					
					%}
     \vspace{-13pt}
					\caption{Efficiency verification experiments on training time, testing time, and memory allocation. }%\kx{Font size, and sugfigure height.}}    % 整个图片的说明，注释写在{}内
					\label{figeffency}           
    \vspace{-15pt}
	\end{figure}
\subsection{Computational Complexity Comparison}
\noindent\textbf{Q: Whether the PGR-MOOD reduces the complexity of time and space in the training and testing phases?}
Yes, to validate the efficiency and scalability of PGR-MOOD, we conduct comprehensive comparisons against the SOTA method GOOD-D and a baseline GR-MOOD. The comparative results are illustrated in Fig.~\ref{figeffency}. 
Although PGR-MOOD slightly trails GOOD-D in testing time, it markedly surpasses it in all other aspects.

% \subsubsection{Efficiency on execution time}
\noindent$\rhd$ \textsf{Efficiency on execution time.} 
During the training phase, PGR-MOOD exhibits a substantially reduced training duration compared to both GOOD-D and GR-MOOD. This efficiency stems from GOOD-D's reliance on a time-consuming contrastive learning approach for model training, whereas GR-MOOD necessitates fine-tuning of the diffusion model on the training set. 
In contrast, PGR-MOOD requires the generation of only a limited set of prototype graphs, thereby enhancing its training efficiency.
During the testing phase, GOOD-D leverages its trained model to directly classify input graphs, while PGR-MOOD's method, which entails calculating the similarity between input graphs and the set of prototypical graphs individually. 
Consequently, PGR-MOOD is marginally slower than GOOD-D. However, it significantly outpaces GR-MOOD, which requires the regeneration of reconstructed graphs for each input. %thus making PGR-MOOD a more efficient alternative in comparison.
% In the training phase, the training time of PGR-MOOD is much smaller than that of GOOD-D and GR-MOOD.The reason for this is that GOOD-D uses a more time-consuming contrast learning method to train the model, while GR-MOOD needs to re-generate a reconstructed graph for each input graph.

% To verify the competitiveness of our method in terms of execution time, we record the training time and test time of PGR-MOOD, SOTA, and GR-MOOD, where PGR-MOOD performs the generation of G during training, SOTA method needs to train a redesigned GNNs.
% GR-MOOD requires training vanilla GNNs and fine-tuning the diffusion model on IDs. As shown in the left subfigure in Fig~\ref{figeffency}, PGR-MOOD requires the shortest training time. In the testing process, SOTA needs to calculate the OOD judge score of the current test graph using GNNs, GR-MOOD needs to reconstruct the original input using the diffusion model, and PGR-MOOD only needs to calculate the similarity between G and the current input. As shown in the middle subfigure in Fig~\ref{figeffency}, GR-MOOD takes the longest time to generate due to the diffusion model, while our method and SOTA have the same test time, but have better performance as mentioned in previous sections.

% \subsubsection{Efficiency on memory allocation}
\noindent$\rhd$ \textsf{Scalability in memory allocation.} 
To assess the memory efficiency of our method, we evaluate memory allocation during the testing phase. PGR-MOOD, which eschews the need for any model for OOD detection, only loads the set of prototypical graphs and demands the least memory allocation.
In contrast, the GOOD-D method requires loading GNNs, and GR-MOOD necessitates loading a diffusion model for reconstruction graphs, thereby increasing their memory requirements. The experimental findings underscore that our approach%, which leverages the prototypical graph $\overline{G}$ as an assembly of reconfiguration graphs for similarity computation, 
can significantly mitigate memory consumption and enhance model scalability.
% In order to verify the efficiency of our method in terms of memory allocation, we count it during test phase. Since PGR-MOOD does not require any model for OOD detection, it requires the least memory allocation. GOOD-D method needs to load a GNNs and GR-MOOD needs to load a diffusion model for reconstruction, so it needs the most memory. The experimental results prove that our proposed $\overline{G}$ as the prototypical graph for the reconstruction results of the test graph is very effective in reducing the memory proportion

\section{conclusion}
This study explores OOD detection for molecular graphs, starting with a basic diffusion model-based approach, GR-MOOD, and identifying key challenges. We introduce PGR-MOOD, an advanced OOD detection method for molecular graphs that addresses GR-MOOD's limitations by using a diffusion model to create prototypical graphs. These graphs closely resemble ID inputs while distinctly diverging from OOD inputs. PGR-MOOD utilizes the Fused Gromov-Wasserstein distance for efficient similarity measurement and OOD scoring, significantly reducing computational load. Our approach demonstrates SOTA results across ten datasets, proving its effectiveness.
% In this paper, we focus on the OOD detection on molecular graphs. We first verify the feasibility of reconstruction strategy based on diffusion model through a naive version GR-MOOD and pose two major challenges. Based on these, we propose a novel OOD detection method for molecular graphs, named PGR-MOOD, which can overcome all shortcomings of GR-MOOD. At the core of our method lies the utilization of a diffusion model to generate prototypical graphs, which closely match the ID inputs while maximize their distance from OOD inputs. 
% The above-mentioned similarity measurement between graphs in PGR-MOOD is based on Fused Gromov-Wasserstein distance (FGW). In the testing phase, the prototypical graphs are compared with test samples to efficiently determine the OOD score, which significantly reduce the computational overhead. The state-of-the-art experimental results on ten datasets demonstrate the superiority of our method.

% PGR-MOOD significantly streamlines the detection process by circumventing the need for reconstructing every test sample.
%PGR-MOOD's effectiveness hinges on three key innovations: (i)A Strong Similarity Function Fused Gromov-Wasserstein distance (FGW); (ii) Informative Prototypical Graphs Reconstruction; and (iii) 
%PGR-MOOD significantly reducing the computational overhead of the testing phase.  Experimental results demonstrate the superiority of our approach.
\bibliographystyle{ACM-Reference-Format}
\bibliography{reference}

% \newpage
\clearpage

\appendix

\section{Appendix}

\subsection{Descriptions of Datasets and Metric}\label{A.1}
\begin{itemize}

\item DrugOOD~\cite{drugood} is a systematic OOD dataset curator and benchmark for drug discovery, providing
large-scale, realistic, and diverse datasets for graph OOD learning problems. To meet this purpose of covering a wide range of shifts that naturally occur in molecular graphs, we cautiously consider three properties as the basis of dividing ID and OOD, including assay, molecular size, and molecular scaffold.
DrugOOD provides an automated method for dividing datasets into ID training sets, ID testing sets, and OOD testing sets. We use the ID training set to generate prototypical graphs during the training phase, and process OOD detection on the ID testing set and OOD testing set since they have different data distributions.
\item 
%Specifically, we randomly select 1000, 300, 500 graphs from the original training set to formulate
%the ID training, validation and testing datasets. Our OOD mixed dataset is consisted of 1000 graphs
%from the original OOD validation dataset. Finally, 500 graphs are drawn from the original OOD
%testing dataset to compose our OOD testing dataset. For debiased learning, we treat the original
%training dataset as ID data, and the original testing dataset as training outliers, to create larger
%distribution shifts.
GOOD~\cite{good} is a systematic graph OOD benchmark, which provide carefully designed data environments for
distribution shifts. Given a domain, it has two kinds of shift strategies: covariate shift, and concept shift. For a supervised dataset, each inputs $X\in\mathcal{X}$ corresponding to outputs $Y\in\mathcal{Y}$ and have the distribution of training set $P^{train}(\cdot)$ and testing set $P^{test}(\cdot)$. The the joint distribution $P(Y,X)$ can be written as $P(Y,X)=P(Y|X)P(X)$. In covariate shift, the input distributions have been shifted
between training and test data. Formally $P^{train}(X)\neq P^{test}(X) $ and $P^{train}(Y|X)= P^{test}(Y|X) $. For concept shift, the conditional distribution $P(Y|X)$ has been shifted as $P^{train}(X)=P^{test}(X) $ and $P^{train}(Y|X)\neq P^{test}(Y|X)$. In order to maintain the consistency of datasets we adopted covariate shift.
\item AUROC (Area Under the Receiver Operating Characteristic curve), AUPR (Area Under the Precision-Recall curve), and FPR95 (False Positive Rate at $95\%$ True Positive Rate) are metrics commonly used to evaluate the performance of classification models, particularly in the context of binary classification and anomaly or outlier detection tasks such as OOD (Out-Of-Distribution) detection.
\end{itemize}

\subsection{Descriptions of Baseline Methods}\label{A.2}
In our experiments, we compare the following six methods as baselines:
\begin{itemize}
\item MSP~\cite{msp}: MSP utilizes the backbones’ max softmax output as the judge score, where ID has the highest score and OOD has the lowest score.
    \item GOOD-D~\cite{Good-d}: By performing hierarchical contrastive learning on the augmented graphs, GOOD detects OOD graphs based on the semantic inconsistency in different granularities.
    \item GraphDE~\cite{graphde}: GraphDE modeling the graph generative process to characterize the distribution shifts of graph data together with an additionally introduced latent environment variable as an indicator to detect OODs.
    \item AAGOD~\cite{AAGOD}: AAGOD proposes a learnable amplifier to increase the focus on the key pattern of the structure to enlarge the difference between IDs and OODs.
    \item OCGIN~\cite{OCGIN}: OCGIN is a graph anomaly detection with a binary classifier where a GIN encoder by the guide of SVDD~\cite{svdd}.
    \item GLocalKD~\cite{GLocalKD}: GLocalKD proposes a deep graph anomaly detector based on knowledge distillation for both local and global graphs.
\end{itemize}

\subsection{Analysis of Hyper-Parameters}\label{A.3}
To analyze the hyper-parameter sensitivity of PGR-MOOD, we experiment on two datasets with different $\alpha$ and $I$.

\begin{figure}[!t]    
					\centering            % 前面说过，图片放置在中间
					\subfloat[]   % 第一张子图的下标（注意：注释要写在[]中括号内）
					{
						\label{apha}\includegraphics[width=0.20\textwidth,height=0.13\textheight]{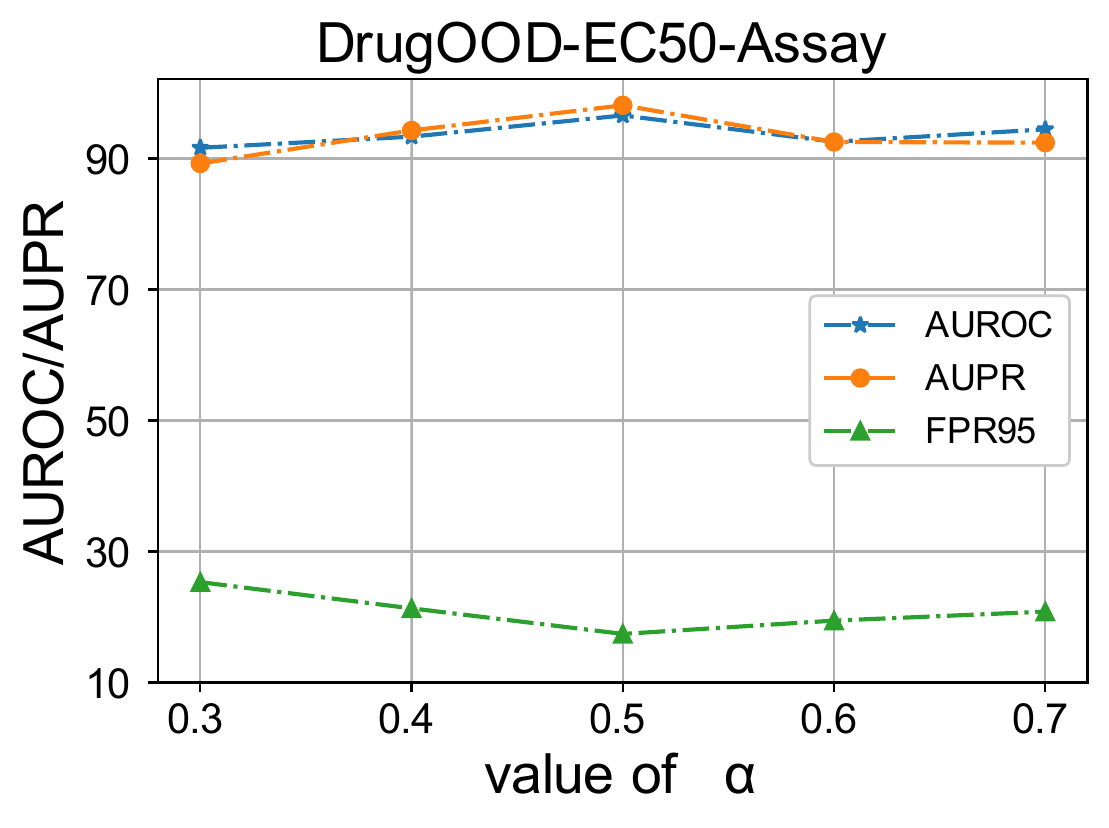}
						
					}\hspace{2mm}
					\subfloat[]
					{
						\label{length}\includegraphics[width=0.20\textwidth,height=0.13\textheight]{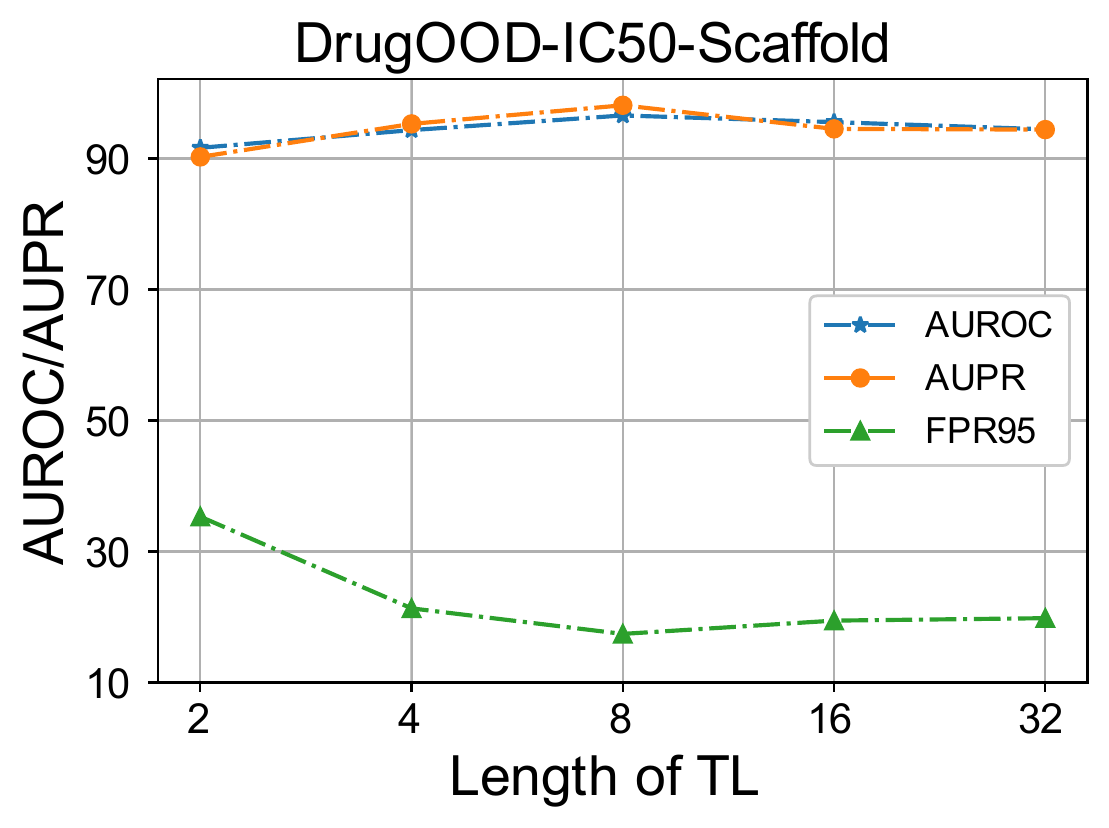}
					}
					
					 \vspace{-10pt}
					\caption{ Analysis of Hyper-Parameters of our method on two DrugOOD datasets. }    % 整个图片的说明，注释写在{}内
					\label{figinfluence}            % 整个图片的标签编号，注意这里跟
     
\end{figure}

\subsubsection{Analysis of $\alpha$}
To analyze the impact of hyper-parameters $\alpha$ in Eq. (\ref{FGW}), which balance the structure term and feature term. 
We vary $\alpha$ in $\{ 0.3, 0.4, 0.5, 0.6, 0.7\}$  and
present the experimental results in Fig. \ref{apha}. PGR-MOOD performs best with $\alpha$ equal to 0.5, which means it is the fairest way for structure and feature. This fits our needs because we can't predict which way the OOD shift will be biased, so it makes sense to weight both terms equally.

\subsubsection{Analysis of $I$}
To analyze the impact of hyper-parameters $I$ in Eq. (\ref{templateOOD}), which corresponds to the number of prototypical graph $\overline{G}$ that we need to generate. 
We vary $I$ in $\{2,4,8,16\}$  and
present the experimental results in Fig. \ref{length}. The performance of PGR-MOOD is stable when $I$ changes. In fact, the size of $I$ does not have a huge impact on the final OOD detection result. The calculation of $\overline{G} \in PL$ can eventually traverse the entire $D_{in}$, only the memory required for the generation process will be affected.

% \begin{figure}[!t]    
% 					\centering            % 前面说过，图片放置在中间
% 					\subfloat[experiment performance on DrugOOD-IC50-Scaffold]   % 第一张子图的下标（注意：注释要写在[]中括号内）
% 					{
% 						\label{apha}\includegraphics[width=0.22\textwidth,height=0.17\textheight]{Ainfluence.pdf}
						
% 					}\hspace{2mm}
% 					\subfloat[experiment performance on DrugOOD-EC50-Assay]
% 					{
% 						\label{length}\includegraphics[width=0.22\textwidth,height=0.17\textheight]{Kinfluence.pdf}
% 					}

% 					\caption{ Our method. }    % 整个图片的说明，注释写在{}内
% 					\label{figinfluence}            % 整个图片的标签编号，注意这里跟
% \end{figure}

\end{document}